%% file: acl_latex.tex
\documentclass[11pt]{article}

\usepackage[final]{acl}

\usepackage{times}
\usepackage{latexsym}

\usepackage[T1]{fontenc}
\usepackage{caption} % for \ContinuedFloat
\newcommand{\reasoning}[1]{{\small #1}}
\usepackage{soul}
\usepackage[table]{xcolor} % optional, lets you customize the highlight color
\sethlcolor{yellow!35} 
\usepackage[utf8]{inputenc}

\usepackage{microtype}

\usepackage{inconsolata}
\usepackage{comment}
\usepackage{graphicx}
\usepackage{enumitem}
\usepackage{amsmath}
\usepackage{booktabs}      % top/mid/bottomrule, cmidrule, specialrule
\usepackage{tabularx}
\usepackage{float}
\usepackage{multirow}      % \multirow
\usepackage[table]{xcolor} % \rowcolor, color in tables
\usepackage{graphicx}      % \resizebox
\usepackage{array}
\newcolumntype{L}{>{\raggedright\arraybackslash}p{0.23\textwidth}}
\newcolumntype{Y}{>{\raggedright\arraybackslash}X}
\usepackage{subcaption}
\usepackage{tikz}
\usepackage[most]{tcolorbox}
\tcbuselibrary{skins, breakable} % needed for enhanced + float support
\tcbset{
  promptstyle/.style={
    enhanced,
    breakable,              % <-- allows page breaks
    width=\linewidth,       % column width in two-column mode
    colback=gray!10,
    colframe=black,
    boxrule=0.5pt,
    arc=4pt,
    left=6pt, right=6pt,
    top=6pt, bottom=6pt,
    fontupper=\ttfamily,
    colbacktitle=black,
    coltitle=white,
    title style={font=\bfseries, halign=flush left},
    boxed title style={arc=4pt},
    attach boxed title to top left={xshift=0mm, yshift=-2mm},
  }
}
\newtcolorbox{prompt}[1][]{promptstyle,title={Prompt},#1}
\newtcolorbox{examplebox}[1][]{
  colback=gray!10,
  colframe=gray!80,
  fonttitle=\bfseries,
  title=Example,
  left=2mm,
  right=2mm,
  top=1mm,
  bottom=1mm,
  #1
}
\newcommand{\zstar}{\rlap{\textsuperscript{*}}}
\usepackage{amsfonts}
\definecolor{oiBlue}{RGB}{0,114,178}      % Okabe–Ito blue
\definecolor{oiVermillion}{RGB}{213,94,0}
\newcommand{\up}[1]{\textcolor{green!60!black}{\(\uparrow\)}#1}

\newif\iftodos
\todostrue
\providecommand{\usercomment}[3]{%
  {\color{#1}\textbf{#2} #3}%
}

\DeclareRobustCommand{\todo}[2][]{%
  \iftodos
    \usercomment{red}{%
      TODO\if\relax\detokenize{#1}\relax\else~(#1)\fi:%
    }{#2}%
  \fi
}

\title{Distill and Align Decomposition for Enhanced Claim Verification}

\author{
 \textbf{Jabez Magomere,\textsuperscript{1,2}} \thanks{Work done during an internship at JPMorgan AI Research.} 
 \textbf{Elena Kochkina,\textsuperscript{2}}
 \textbf{Samuel Mensah,\textsuperscript{2}} 
 \textbf{Simerjot Kaur,\textsuperscript{2}} 
 \textbf{Fernando Acero,\textsuperscript{2}} \\
 \textbf{Arturo Oncevay,\textsuperscript{2}}
 \textbf{Charese H. Smiley,\textsuperscript{2}}
 \textbf{Xiaomo Liu,\textsuperscript{2}}
 \textbf{Manuela Veloso\textsuperscript{2}}
 \\
 \textsuperscript{1}University of Oxford,
 \textsuperscript{2}JPMorgan AI Research
 \\
 \texttt{jabez.magomere@keble.ox.ac.uk} 
 \\
  \texttt{\{name\}.\{surname\}@jpmorgan.com} 
 }

\begin{document}
\maketitle

\begin{abstract}

Complex claim verification requires decomposing sentences into verifiable subclaims, yet existing methods struggle to align decomposition quality with verification performance. 
We propose a reinforcement learning (RL) approach that jointly optimizes decomposition quality and verifier alignment using Group Relative Policy Optimization (GRPO). Our method integrates: (i) structured sequential reasoning; 
(ii) supervised finetuning on teacher-distilled exemplars; and (iii) a multi-objective reward balancing format compliance, verifier alignment, and decomposition quality. 
Across six evaluation settings, our trained 8B decomposer improves downstream verification performance to \(71.75\%\) macro-F1, outperforming prompt-based approaches (\(+1.99\), \(+6.24\)) and existing RL methods (\(+5.84\)). 
Human evaluation confirms the high quality of the generated subclaims.
Our framework enables smaller language models to achieve state-of-the-art claim verification by jointly optimising for verification accuracy and decomposition quality.
\end{abstract}

\section{Introduction}
\input{sections/introduction}

\section{Problem Formulation}
\input{sections/background}

\label{sec:background}
\section{Our Approach}
\label{sec:methods}
\input{sections/method}

\section{Experimental Setup}
\label{sec:experiments}
\input{sections/experimental_setup}
\section{Results \& Discussion}
\input{sections/results}
\section{Decomposition Quality Estimation}
\input{sections/quality_estimation}
\section{Related Works}
\input{sections/related_works}
\section{Conclusion}
\input{sections/conclusion}

\section*{Limitations}
\input{sections/limitations}

\section*{Ethical Considerations}
Our work uses publicly available datasets and does not involve private or sensitive data. However, factuality and neutrality of sources like Wikipedia and web search are not guaranteed, and any biases or inaccuracies may affect system outputs. Automated fact verification is not infallible; errors in decomposition or verification could lead to incorrect judgments, especially in sensitive domains. Our models may also inherit social biases from training data, and we do not explicitly address bias mitigation.  Human evaluation was conducted by expert annotators following clear guidelines; annotators were blinded to system identity to reduce bias, and no personally identifiable information was involved. Finally, our experiments focus on English-language data, and ethical implications may differ in other languages or cultural contexts.

\section*{Acknowledgments}
We thank Uljad Berdica, Caiqi Zhang, Salim Amoukou and Myeong-jun Erik Jang for insightful discussions and valuable feedback. 

\paragraph{Disclaimer}This paper was prepared for informational purposes by the Artificial Intelligence Research group of JPMorgan Chase \& Co. and its affiliates ``JP Morgan'') and is not a product of the Research Department of JP Morgan. JP Morgan makes no representation and warranty whatsoever and disclaims all liability, for the completeness, accuracy or reliability of the information contained herein. This document is not intended as investment research or investment advice, or a recommendation, offer or solicitation for the purchase or sale of any security, financial instrument, financial product or service, or to be used in any way for evaluating the merits of participating in any transaction, and shall not constitute a solicitation under any jurisdiction or to any person, if such solicitation under such jurisdiction or to such person would be unlawful. 
%© 2026 JPMorgan Chase \& Co. All rights reserved 

\bibliography{custom}

%\newpage
\appendix

\section{Annotation Guidelines}
\label{sec:appendixA}
\input{sections/appendixA}

%This is an appendix.
\clearpage
\onecolumn
\input{sections/prompts}

\end{document}

%% file: sections/introduction.tex
Verifying the factual accuracy of complex long-form text generated by LLMs has become critical as these systems are increasingly deployed in high-stakes applications~\cite{augenstein2024factuality, li-etal-2024-dawn, bang-etal-2025-hallulens}. The \textit{decompose-then-verify} pipeline has emerged as the dominant framework for this task: model responses are split into subclaims, each subclaim is verified against retrieved evidence, and verification results are aggregated to determine overall factuality~\citep{min_factscore_2023, wei_long-form_2024, song_veriscore_2024}. 

\begin{figure*}[t]
  \centering
  \includegraphics[width=1\textwidth]{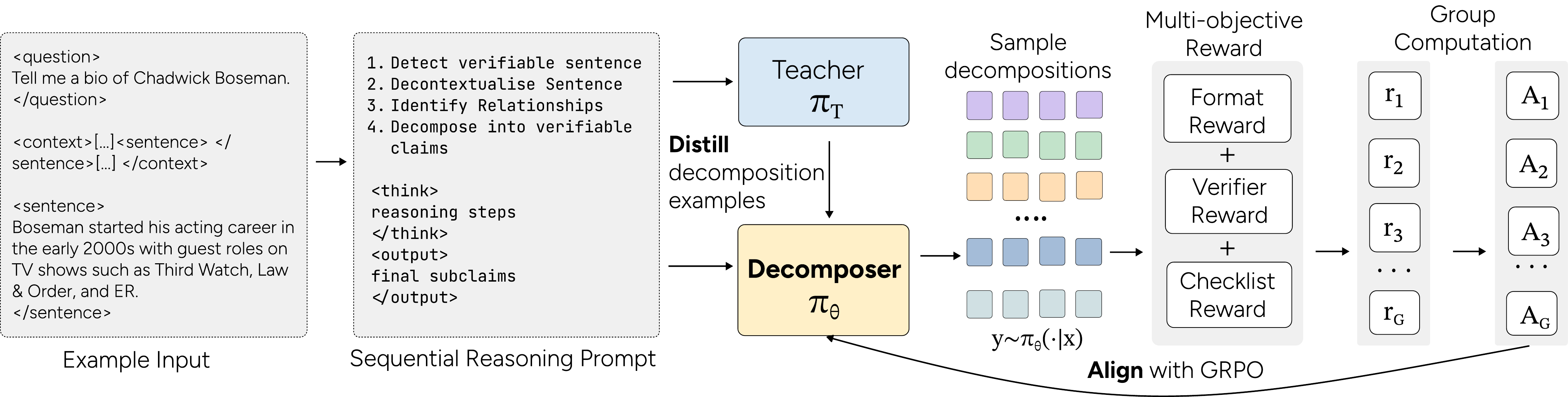}
    \caption{\textbf{Distill-Align-Decompose (DAD).} We (1) distill examples from a teacher model, (2) frame decomposition as a sequential reasoning task, and (3) align the policy via GRPO using multi-objective rewards.}
  \label{fig:diagram}
\end{figure*}
However, the decomposer can introduce noise such as over-fragmentation, context loss, or insufficient decontextualisation—which harms retrieval and downstream verification \citep{hu-etal-2025-decomposition, liu2025verifactenhancinglongformfactuality}. Additionally, decomposers and verifiers are typically optimised independently~\cite{seo2025verifying}, producing subclaims at an atomicity level\footnote{We follow the atomicity definition from \citet{lu_optimizing_2025}, a metric quantifying information density, atomicity = $log_2$(\# atomic information), where one piece of atomic information is an utterance conveying a single nontrivial fact. Higher atomicity means a claim is more coarse-grained and information-rich.} the verifier may not be trained to handle~\citep{lu_optimizing_2025}. Balancing decomposition quality with verifier alignment is difficult: the verifier’s preferred atomicity is latent, and decomposition quality often lacks a reliable, objective metric. Prior work addresses them separately, improving quality with complex multi-stage pipelines that add cost and latency \citep{liu2025verifactenhancinglongformfactuality, metropolitansky_towards_2025}, or improving verifier alignment by deciding only when to decompose without improving the decomposition quality \citep{lu_optimizing_2025}. We argue that these limitations stem from a fundamental issue: current methods lack a unified training objective that jointly optimizes quality and alignment while remaining computationally efficient.

To address this, we propose a training recipe (shown in Figure~\ref{fig:diagram}) with three components. (1) We reframe decomposition as a \emph{sequential reasoning task}, requiring the model to explicitly reason through intermediate steps before generating subclaims. (2) We initialize a decomposer policy via supervised fine-tuning on synthetic decompositions distilled from a larger teacher model. (3) We train the decomposer with GRPO~\cite{shao2024deepseekmathpushinglimitsmathematical} using a multi-objective reward that combines verifier feedback for alignment with an explicit decomposition quality checklist.

Across six evaluation settings, our trained 8B decomposer improves downstream verification performance to \(71.75\%\) macro-F1, outperforming prompt-based approaches (VeriScore~\cite{song_veriscore_2024}: \(+1.99\)pp), existing RL methods (DyDecomp~\cite{lu_optimizing_2025}: \(+5.84\)pp), and its own prompt-only variant (\(+2.58\)pp). Our method generates \(8.14\) subclaims on average, avoiding both FActScore's~\cite{kamoi_wice_2023} overdecomposition (\(22.92\) subclaims) and DyDecomp's underdecomposition (\(1.66\) subclaims), making the verification process more efficient. Human evaluation confirms high scores across all quality desiderata, validating that our multi-objective training improves both verification accuracy and decomposition quality. We make the following contributions \footnote{Code will be made available upon request.}:
\begin{itemize}[leftmargin=*, noitemsep]
    \item We propose a sequential reasoning framework that structures decomposition in a single model call, improving quality without added latency.
    \item We introduce a  multi-objective GRPO training approach that jointly optimizes verifier alignment and decomposition quality, demonstrating these objectives need not be traded off.
    \item We conduct human evaluation demonstrating high quality across all desiderata, validating that our approach improves both verification accuracy and decomposition quality.
\end{itemize}

%% file: sections/background.tex
\label{sec:problem_formulation}
The \textit{decompose-then-verify} pipeline is a post-hoc factuality evaluation framework for verifying long-form model responses \citep{min_factscore_2023, kamoi_wice_2023, wei_long-form_2024, song_veriscore_2024, wanner-etal-2024-closer}. In this framework, each sentence in a response is passed to a decomposer LLM $\mathcal{D}$, whose goal is to generate a set of verifiable subclaims $\{c_1, c_2, \ldots, c_n\}$. Each subclaim $c_i$ is then processed by an evidence retrieval module $\mathcal{R}$, which queries a knowledge source (e.g., Wikipedia or web search API) and returns a set of evidence snippets $E_i = \{e_{i1}, \ldots, e_{ik_i}\}$. 
The verifier module $\mathcal{V}$ evaluates each claim together with its retrieved evidence and outputs factuality label $y_i$, where $y_i \in \{\textsc{Supported}, \textsc{Not Supported}\}$. The set of claim-level predictions $\{y_i\}$ is then aggregated to produce an overall factuality judgment. 

The central challenge in decomposition for accurate factuality verification is to ensure that the generated subclaims are both high quality and aligned with the downstream verifier. Prior work has proposed several desiderata for decomposition quality. Adapting these to the verification setting, we define the task formally: given a sentence $S$, a decomposer $\mathcal{D}$ generates a set of subclaims $C = \{c_1, c_2, \ldots, c_n\}$, which should satisfy the following properties:

\begin{enumerate}[noitemsep, topsep=0pt]
    \item \textbf{Atomicity Alignment}: Each subclaim $c_i$ is decomposed to the level of granularity expected by the downstream verifier $\mathcal{V}$ \citep{lu_optimizing_2025}.
    \item \textbf{Verifiability}: Each subclaim $c_i$ constitutes a verifiable proposition, i.e., ``a statement or assertion that can be objectively verified as true or false based on empirical evidence or reality,'' for which the verifier can assign a label $y_i \in \{\textsc{Supported}, \textsc{Not Supported}\}$ \citep{song_veriscore_2024}.
    \item \textbf{Entailment}: Each subclaim $c_i$ is entailed by the original sentence ($S \models c_i$) without introducing spurious information \cite{wanner-etal-2024-closer}.
    \item \textbf{Coverage}: The set of subclaims $\{c_i\}$ collectively captures all verifiable facts expressed in $S$ \citep{hu-etal-2025-decomposition, metropolitansky_towards_2025}.
    \item \textbf{Decontextualization}: Each subclaim $c_i$ is interpretable in isolation, with entities specified and pronouns resolved \citep{gunjal_molecular_2024}.
\end{enumerate}

Existing approaches fail to jointly meet these criteria, leaving the challenge of producing subclaims that are both high quality and verifier-aligned. Our proposed approach aims to address this challenge.

%% file: sections/method.tex
To jointly ensure decomposition quality and verifier alignment, we propose a training recipe with three components presented next.

\subsection{Reframing Decomposition as a Sequential Reasoning Task}
We reframe decomposition as a step-by-step sequential reasoning task executed within a \emph{single model call per sentence}. 
Each step conditions the next and requires the decomposer to articulate its reasoning before producing subclaims. 
This design enforces explicit reasoning about verifiability, decontextualization, relationships, and claim boundaries, while remaining computationally efficient.

Given a model response segmented into sentences ${S_1,\ldots,S_n}$, we define for each $i \in [n]$ a decomposition input $x_i = (Q, C_i, S_i)$, where $Q$ is the original question (or prompt) corresponding to the model response, and $C_i$ denotes the local context of $S_i$, consisting of the $p$ preceding and $f$ following sentences, truncated to $[1,n]$. For each target sentence $S_i$, we construct a local context window including the $p=2$ preceding and $f=2$ following sentences. This matches the context window size used by \citet{song_veriscore_2024}, ensuring a fair comparison, but adopts a symmetric split rather than VeriScore’s \cite{song_veriscore_2024} asymmetric split (3 preceding, 1 following) in order to equally capture both antecedents and consequences of a sentence. The decomposer $\mathcal{D}$ takes as input the triple $x_i = (Q, C_i, S_i)$ and applies a four-step reasoning process below.\footnote{The system prompt used by $\mathcal{D}$ is provided in Appendix~\ref{appendix:prompts}.}

\begin{enumerate}[noitemsep, topsep=0pt]
    \item \textbf{Claim detection} ($S_i \rightarrow$ VerifiableSentence): Determine whether $S_i$ contains at least one verifiable proposition (i.e., a fact that can be checked against evidence). If so, return the portion containing verifiable information; otherwise return \textsc{No Verifiable Claim}.
    \item \textbf{Decontextualization} (VerifiableSentence $\rightarrow$ DecontextualizedSentence): Rewrite the sentence so it is self-contained and unambiguous, resolving references using only $Q$ and $C_i$. If this is not possible, return \textsc{Cannot be Decontextualized}.
    \item \textbf{Relationship identification} (DecontextualizedSentence $\rightarrow$ [Relationships]): Identify any logical or discourse relations (e.g., attribution, causal, temporal, comparison, expansion, negation, membership) necessary to preserve the meaning of the original sentence.
    \item \textbf{Claim extraction} (DecontextualizedSentence + Relationships $\rightarrow$ Claims): Split into minimal factual units, each fully decontextualized, verifiable, and interpretable in isolation, while preserving identified relationships.
\end{enumerate}
The decomposer is instructed to return its output $y$ consisting of reasoning steps inside \texttt{<think>} tags, and the subclaims inside \texttt{<output>} tags.

\subsection{Distilling High-Quality Decomposition Exemplars from a Teacher Model}
To mitigate cold-start issues before applying RL, we first distill decomposition exemplars from a larger teacher policy $\pi_T$ and use them to initialize a student policy $\pi_\theta$. This warm-start serves two purposes: (i) the student acquires instruction-following and output formatting skills prior to RL training, and (ii) the student acquires basic decomposition capabilities from the stronger teacher. The student policy $\pi_\theta$ is trained with a standard supervised fine-tuning (SFT) objective, minimizing the token-level negative log-likelihood of the teacher-generated decomposition examples.

\subsection{Aligning the Student Policy via Reinforcement Learning with a Multi-Objective Reward}
% We now introduce the reinforcement learning stage of our proposed framework.
\subsubsection{Reinforcement Learning Formulation}
We formulate the problem of aligning the student decomposer $\pi_\theta$ as a single-step Markov Decision Process (MDP). Formally, the state space $x \in \mathcal{S}$ corresponds to the input $x=(Q, C_i, S_i)$, and the action space $y \in\mathcal{A}$ consists of all possible decomposition outputs $y$ (a reasoning trace followed by subclaims). The student policy is parameterized as $\pi_\theta(y\mid x)$, and the reward function $R(x,y)$ is a multi-objective signal combining format, verifier, and checklist reward terms. Each episode is single-step: the agent observes an input $x$, produces an output $y \sim \pi_\theta(\cdot \mid x)$, and receives the corresponding reward $R(x,y)$. The student policy $\pi_\theta$ is fine-tuned via reinforcement learning to maximize the expected reward $\mathbb{E}_{x \sim \mathcal{D}, y \sim \pi_\theta(\cdot \mid x)}[R(x,y)]$ under this formulation.

\subsubsection{Group Relative Policy Optimization}
\label{sec:grpo}
We optimize our policy using Group Relative Policy Optimization (GRPO) \citep{shao2024deepseekmathpushinglimitsmathematical}. GRPO samples multiple outputs per input and computes group-relative advantages by comparing each sample’s reward against the mean reward of its group, which yields more stable gradients when reward distributions vary across input complexities. GRPO also eliminates the need for a separate critic network, making it computationally efficient for fine-tuning large language models.

\begin{equation}
\label{eq:grpo}
\begin{aligned}
\mathcal{L}_{\text{GRPO}}(\theta)
&= \mathbb{E}_{x \sim \mathcal{D},\, y \sim \pi_\theta(\cdot \mid x)} \Big[ \\
&\quad \frac{\pi_\theta(y \mid x)}{\pi_{\text{ref}}(y \mid x)}
\cdot \hat{A}_{\text{group}}(x,y)
\Big] ,
\end{aligned}
\end{equation}
where $\hat{A}_{\text{group}}(x,y)$ is the group-relative advantage defined as
\begin{equation}
\hat{A}_{\text{group}}(x,y)
= R(x,y) - \frac{1}{|B|} \sum_{(x',y') \in B} R(x',y') ,
\end{equation}
with $B$ denoting a batch of sampled input--output pairs and $R(x,y)$ the reward function.

\subsubsection{Multi-Objective Reward Function}
To align the policy $\pi_\theta$ with our decomposition desiderata and the downstream verifier, we design a multi-objective reward function comprising three reward terms: a format reward for structured output, a verifier reward to capture atomicity preferences, and a checklist reward for subclaim quality. During training, GRPO samples completions $y_1, \ldots, y_k \sim \pi_\theta(\cdot|x)$ for each input $x$ and scores them with these reward terms to compute group-relative advantages. Each reward term is normalized to $[0,1]$, defined as follows:

\noindent\textbf{Format Reward.} Correct formatting is essential for downstream parsing: reasoning steps must appear in \texttt{<think></think>} tags and subclaims in \texttt{<output>[]</output>} tags as a valid list. We use a soft reward with partial credit (rather than binary pass/fail) to provide more informative gradients for the policy. The format reward is computed as $r_{\text{format}}(y) = \sum_{i} w_i \cdot f_i(y)$, where $f_i(y) \in \{0,1\}$ are binary checks for: presence of both required tags, correct tag ordering, successful list parsing, and non-empty list with no blank strings, and $w_i$ is a weight assigned to each check (Appendix ~\ref{appendix:reward-implementation}). We use a soft format reward instead of structured generation to maintain compatibility with RL training, reduce training overhead, and draw on prior evidence that soft rewards provide richer learning signals \citep{damani2025binaryrewardstraininglms, shao2024deepseekmathpushinglimitsmathematical}.          

\noindent\textbf{Verifier Reward.} Since the verifier's preferred atomicity level cannot be directly modeled, we use verifier prediction accuracy as a proxy: outputs that yield accurate verifier predictions reveal the appropriate subclaim atomicity for a given verifier. For each output, we extract subclaims and retrieve $k$ supporting evidence snippets from a knowledge source for each subclaim $c_i$. A verifier model $\mathcal{V}$ scores each claim–evidence pair to obtain factuality probabilities $p_i \in [0,1]$, which are aggregated into a sentence-level prediction and compared to the ground-truth label $\ell \in \{0,1\}$ of sentence $S_i$ from the input. We compare two verifier reward designs:
\begin{enumerate}[nosep]
    \item \textbf{Sparse.} Aggregate claims via a logical AND: predict \textsc{Supported} if and only if all $p_i \ge \tau$. Let $\hat{\ell} \in \{0,1\}$ denote the resulting sentence-level prediction. The sparse verifier reward is $r_{\text{ver}}(x,y) \in \{0,1\}$, where $r_{\text{ver}}(x,y) = 1$ if $\hat{\ell} = \ell$ and 0 otherwise.
    \item \textbf{Dense.} Following \citet{damani2025binaryrewardstraininglms}, we replace binary accuracy with the Brier score: for predicted probability $q\in[0,1]$ and outcome $o\in\{0,1\}$, $\mathrm{Brier}(q,o)=(q-o)^2$. For a list of subclaims with verifier probabilities $p_1,\dots,p_m$, we compute the sentence-level probability $\bar{p}$ via the geometric mean, which is more sensitive to low confidences. The dense verifier reward is $r_{\text{ver}}(x,y)=1-(\bar{p}-\ell)^2$.
\end{enumerate}
\noindent\textbf{Checklist Reward.} 
To ensure high-quality decompositions, we introduce a weighted checklist rubric that evaluates each decomposed subclaim against multiple criteria. We employ an LLM-as-a-Judge $\mathcal{J}$, which assesses whether each criterion $c_i$ is satisfied for each subclaim. The judge outputs a binary score for each criterion, $\mathcal{J}(x, y, c_i) \rightarrow \{0, 1\}$, and the checklist reward for each subclaim is calculated as the weighted average $\frac{\sum_{i} w_i \cdot \mathcal{J}(x, y, c_i)}{\sum_{i} w_i}$. The final checklist reward $r_{\text{checklist}}(x,y)$ is the geometric mean across all subclaims. The criteria cover: complete verifiability (single verifiable proposition), retrieval relevance with respect to the question, presence of all necessary qualifiers, explicit references to entities and relations, and absence of ungrounded additions. The complete checklist with detailed criteria and prompts is in Appendix~\ref{appendix:reward-implementation}.

\noindent\textbf{Overall Reward.}
For an input $x \sim \mathcal{D}$ and completion $y \sim \pi_\theta(\cdot \mid x)$, the reward function is the equally-weighted sum of our three reward terms:
\[
R(x,y) = r_{\text{format}}(y) + r_{\text{ver}}(x,y) + r_{\text{checklist}}(x,y).
\]

%% file: sections/experimental_setup.tex
\subsection{Datasets}
\noindent\textbf{SFT.} We build the SFT dataset from open-model responses on long-form factuality benchmarks (VeriScore~\cite{song_veriscore_2024} and VeriFastScore~\cite{rajendhran2025verifastscorespeedinglongformfactuality}). We sample a balanced dataset of \textbf{15.7k} sentences with surrounding context from these responses, varying prompt source, model source, and sentence lengths. We then use the teacher model with our decomposition prompt to generate synthetic decompositions. After filtering malformed outputs, we obtain \textbf{13.7k} training examples and \textbf{1.5k} test examples.

\paragraph{RL.} For RL training, we require a dataset containing sentence-level factuality labels to compute the verifier reward. We use the dataset constructed by \citet{lu_optimizing_2025}, that contains factuality-annotated sentences across varying input atomicity for responses from ChatGPT and PerplexityAI to FActScore~\cite{min_factscore_2023}. To align with \citet{lu_optimizing_2025} and expose the policy to more complex claims, we retain \emph{Atomicity~1} (single-sentence units) and \emph{Atomicity~2} (multi-sentence spans), producing \textbf{2.3k} training instances. We use the Wikipedia dump from \citet{min_factscore_2023} as the knowledge source.

\paragraph{Evaluation Datasets.} We evaluate across four datasets under six evaluation settings that vary in input granularity, domain, and knowledge source (Table~\ref{tab:evaluation-datasets}). We group settings by annotation granularity: (1) \emph{sentence-level} datasets with \textsc{Supported}/\textsc{Not Supported} labels for individual claim units, and (2) \emph{response-level} datasets with a factuality label for the entire response. For sentence-level evaluation, we use test split from \citet{lu_optimizing_2025}, consisting of \textbf{ChatGPT} and \textbf{PerplexityAI} responses to FActScore biography prompts~\cite{min_factscore_2023}. We evaluate two input granularities for each source: \emph{Atomicity-1} (single sentences) and \emph{Atomicity-2} (multi-sentence spans). %, yielding four sentence-level evaluation settings. 
Evidence retrieval uses top-5 passages from Wikipedia. For response-level evaluation, we use two datasets with Google Search as knowledge source. \textbf{FELM}~\cite{chen2023felm} contains annotated ChatGPT responses across multiple domains, focusing on world knowledge questions. \textbf{BINGCHAT}~\cite{li-etal-2024-self} contains factuality annotations for Microsoft Copilot %(formerly BingChat) 
responses spanning diverse topics; these responses are significantly longer than other datasets. For both datasets, we retrieve top-10 results from Google Search via SerpAPI\footnote{\url{https://www.serpapi.com}} as evidence.

\subsection{Models}
We conduct all experiments with the Llama family of models~\cite{grattafiori2024llama3herdmodels}. 
For the \textbf{decomposer model}, we use \textbf{Llama-3.1-8B-Instruct}.  
For the \textbf{verifier}, we use \textbf{Bespoke-MiniCheck-7B}~\cite{tang_minicheck_2024}\footnote{\url{https://huggingface.co/bespokelabs/Bespoke-MiniCheck-7B}}, a specialised fact-verification model that achieves state-of-the-art performance on aggregated fact-checking benchmarks while being computationally efficient, making it suitable for reward modelling tasks. As the \textbf{teacher model}, we use \textbf{Llama-3.1-405B-Instruct} to generate the synthetic data for the SFT warm-up. For the \textbf{LLM-as-a-Judge}, we use \textbf{Llama-3.3-70B-Instruct}, which provides reliable evaluations at a favourable quality–cost trade-off.

\input{tables/main_results}

\paragraph{Baselines.} We compare our trained decomposer to prompt-based, SFT-based, and RL-based baselines, and probe model-scale effects on verification accuracy. For \textbf{prompt baselines} using Llama-3.1-8B-Instruct, we compare against two few-shot approaches; \textbf{FActScore}~\cite{min_factscore_2023} (extracts atomic facts) and \textbf{VeriScore}~\cite{song_veriscore_2024} (focuses on verifiability), and our decomposition prompt applied zero-shot to ablate training gains. For the \textbf{SFT baseline}, we compare against the open-source VeriScore claim extractor~\cite{song_veriscore_2024} (Mistral-7B), trained on 13.4k GPT-4–distilled decompositions. For the \textbf{RL baseline}, we implement DyDecomp~\cite{lu_optimizing_2025}, a PPO-based approach that trains a policy to dynamically determine when to decompose claims using verifier confidence as a reward. We preserve their configuration (Llama-3-70B-Instruct decomposer, Llama-3-8B-Instruct verifier) for faithful comparison.  
Unlike all other baselines which segment responses into sentences before decomposition, DyDecomp operates directly on full responses. 
To probe model scale effects, we run our zero-shot prompt on Llama-3.3-70B-Instruct and Llama-3.1-405B-Instruct. Implementation details (training and hyper-parameters) of our experiments are provided in Appendix~\ref{appendix:implementation_details}.

\paragraph{Metrics.} For sentence-level datasets, we follow \citet{lu_optimizing_2025}: a claim is \textsc{Supported} if and only if all of its subclaims are \textsc{Supported} (logical AND). For response-level datasets, following \citet{hu-etal-2025-decomposition}, we aggregate subclaim-level verifier scores via harmonic mean; a response is \textsc{Supported} when the aggregate exceeds 0.5. We report  balanced accuracy (BAcc) amd macro-F1 over \{\textsc{Supported}, \textsc{Not Supported}\} to handle class imbalance. We also report the average subclaim count per input to quantify downstream computational cost: each subclaim requires independent evidence retrieval and verifier inference, so higher subclaim counts translate directly to increased latency and compute requirements.

%% file: tables/main_results.tex
\begin{table*}[h]
\centering
\resizebox{\textwidth}{!}{%
\begin{tabular}{ll
ccc ccc  % Perplexity A1 (BAcc,F1,SC) + A2
ccc ccc  % ChatGPT A1 + A2
ccc      % FELM
ccc      % BINGCHAT
ccc      % Overall
}
\toprule
\multirow{3}{*}{\textbf{Decomposer}} & \multirow{3}{*}{\textbf{Approach}}
& \multicolumn{6}{c}{\textbf{PerplexityAI}}
& \multicolumn{6}{c}{\textbf{ChatGPT}}
& \multicolumn{3}{c}{\multirow{4}{*}{\textbf{FELM}}}
& \multicolumn{3}{c}{\multirow{4}{*}{\textbf{BINGCHAT}}}
& \multicolumn{3}{c}{\multirow{4}{*}{\textbf{Overall}}} \\
\cmidrule(r){3-8}\cmidrule(r){9-14}
& &
\multicolumn{3}{c}{\textbf{Atomicity-1}} & \multicolumn{3}{c}{\textbf{Atomicity-2}}
& \multicolumn{3}{c}{\textbf{Atomicity-1}} & \multicolumn{3}{c}{\textbf{Atomicity-2}}
& \multicolumn{3}{c}{} & \multicolumn{3}{c}{} & \multicolumn{3}{c}{} \\
\cmidrule(r){3-5}\cmidrule(r){6-8}\cmidrule(r){9-11}\cmidrule(r){12-14}\cmidrule(r){15-17}\cmidrule(r){18-20}\cmidrule(r){21-23}
& & \textbf{BAcc} & \textbf{F1} & \textbf{SC} & \textbf{BAcc} & \textbf{F1} & \textbf{SC}
  & \textbf{BAcc} & \textbf{F1} & \textbf{SC} & \textbf{BAcc} & \textbf{F1} & \textbf{SC}
  & \textbf{BAcc} & \textbf{F1} & \textbf{SC}
  & \textbf{BAcc} & \textbf{F1} & \textbf{SC}
  & \textbf{BAcc} & \textbf{F1} & \textbf{SC} \\
\addlinespace[0.2em]
\midrule
\rowcolor{gray!10}
\multicolumn{23}{c}{\textit{Baselines}}\\
\cmidrule(l){1-23}
Llama-3.1-8B   & FActScore
& 76.10 & 65.72 & 10.76 & 71.40 & 67.07 & 16.66
& 72.60 & 73.68 & 11.08 & 61.60 & 64.33 & 17.55
& 53.93\zstar & 64.04\zstar & 8.9\zstar
& 50.24\zstar & 58.24\zstar & 72.56\zstar
& 64.31 & 65.51 & 22.92 \\
Mistral-7B     & VeriScore-SFT
& 73.50 & 65.61 & 2.50  & 66.60 & 63.63 & 4.46
& 77.40 & 77.69 & 2.68  & 76.90 & 75.79 & 4.90
& 52.59 & 66.40 & 4.39
& 51.51 & 64.23 & 21.26
& 66.42 & 68.89 & 6.70  \\
Llama-3.1-8B   & VeriScore
& 78.60 & 71.61 & 2.49  & 71.00 & 68.48 & 4.20
& 82.30 & 82.76 & 2.85  & 73.70 & 75.84 & 5.76
& \textbf{59.56\zstar} & 67.86\zstar & 6.65\zstar
& 52.15\zstar & 52.02\zstar & 28.00\zstar
& 69.55 & 69.76 & 8.33  \\
Llama-3-70B    & DyDecomp
& 67.52 & 67.33 & 1.62 & 60.53 & 60.23 & 1.66
& 76.38 & 72.57 & 1.59 & 77.42 & 59.17 & 1.64
& 55.28 & 68.55 & 1.73
& 50.86 & \textbf{67.60} & 1.72
& 64.67 & 65.91 & 1.66  \\
\midrule
\rowcolor{gray!10}
\multicolumn{23}{c}{\textit{DAD Prompt Variants}}\\
\cmidrule(l){1-23}
Llama-3.1-8B   & Prompt
& 73.20 & 67.46 & 2.00 & 69.70 & 68.23 & 3.15
& 79.73 & 79.73 & 2.09 & 75.90 & 71.04 & 3.50
& 56.28 & \textbf{68.57} & 5.05
& \textbf{56.28} & 60.00 & 30.73
& 68.52 & 69.17 & 7.75  \\
Llama-3.3-70B  & Prompt
& 76.60 & 70.45 & 2.47 & 72.90 & \textbf{71.13} & 4.22
& \textbf{84.70} & \textbf{85.01} & 2.60 & 77.90 & 77.93 & 4.73
& 53.38 & 65.81 & 5.06
& 53.38 & 57.48 & 30.10
& 69.81 & 71.30 & 8.20  \\
Llama-3.1-405B & Prompt
& 76.80 & 69.61 & 2.33 & 71.30 & 68.51 & 4.20
& 82.00 & 82.13 & 2.27 & 82.70 & 77.27 & 4.71
& 55.10 & 67.23 & 5.16
& 55.10 & 61.67 & 33.59
& 70.50 & 71.07 & 8.71  \\
\specialrule{.1em}{.25em}{.25em}
\rowcolor{gray!10}
\multicolumn{23}{c}{\textit{Ours}}\\
\cmidrule(l){1-23}
\rowcolor{blue!5}
Llama-3.1-8B & DAD (Ours)
& \textbf{79.70} & \textbf{73.15} & 2.36
& \textbf{73.30} & 70.59 & 3.87
& 81.70 & 81.74 & 2.41
& \textbf{81.60} & \textbf{78.55} & 4.26
& 56.37 & 67.23 & 5.22
& 55.53 & 59.21 & 30.73
& \textbf{71.37} & \textbf{71.75} & 8.14 \\
% \rowcolor{blue!5}
% \multicolumn{2}{l}{(\(\Delta\) vs.\ Prompt Only)}
% & \up{+6.50} & \up{+5.69} &  &
% \up{+3.60} & \up{+2.37} &  &
% \up{+1.97} & \up{+2.01} &  &
% \up{+5.70} & \up{+7.51} &  &
% \up{+2.44} & \up{+1.70} &  &
% \down{-0.75} & \down{-0.79} &  &
% \up{+3.24} & \up{+3.08} &  \\
% \rowcolor{blue!5}
% \multicolumn{2}{l}{(\(\Delta\) vs.\ Best Baseline: VeriScore)}
% & \up{+1.10} & \up{+1.55} &  &
% \up{+2.30} & \up{+2.12} &  &
% \down{-0.60} & \down{-1.02} &  &
% \up{+7.90} & \up{+2.71} &  &
% \down{-0.84} & \up{+3.38} &  &
% \up{+2.41} & \up{+7.19} &  &
% \up{+2.21} & \up{+2.49} &  \\
\bottomrule
\end{tabular}
}
\caption{Balanced Accuracy (BAcc), macro-F1, and subclaim count (SC) per dataset.
Best BAcc and F1 in each column are \textbf{bold}. The final block reports overall BAcc/F1 and the average SC.
Entries marked \(\zstar\) are reported from \citet{hu-etal-2025-decomposition} under a similar setup.
\textbf{On average across six evaluation settings, our 8B decomposer leads to the highest verification performance, with BAcc $71.37\%$ and macro-F1 $71.75\%$, while maintaining comparable granularity (avg.\ $8.14$ subclaims).}}
\label{tab:main_result}
\end{table*}

%% file: sections/results.tex
We present our results in Table~\ref{tab:main_result}, evaluating decomposition approaches across six settings varying input granularity, knowledge sources, and domains. 

\paragraph{Overall performance.} Our Llama-3.1-8B DAD decomposer achieves the highest overall verification performance, with $71.37\%$ balanced accuracy and $71.75\%$ macro-F1 averaged across six evaluation settings. DAD achieves the best performance on three settings (PerplexityAI Atomicity-1, PerplexityAI Atomicity-2 BAcc, and ChatGPT Atomicity-2); larger prompted models perform best on ChatGPT Atomicity-1, and VeriScore or prompt baselines lead on FELM and BINGCHAT. These benchmarks differ substantially in domain, response length, and evidence sources, and prior work shows that decomposition methods often behave inconsistently across datasets due to varying evidence granularity requirements~\citep{hu-etal-2025-decomposition}. Despite this variability, DAD is the only decomposer that achieves the best aggregate performance across all settings. The largest gains over prompt-only baselines occur on ChatGPT Atomicity-2 (\up{+7.51} F1) and PerplexityAI Atomicity-1 (\up{+5.69} F1), showing that specialized training yields meaningful improvements for decomposition-based verification. While absolute gains vary by dataset, reflecting differences in claim atomicity, evidence characteristics, and retrieval difficulty, DAD improves overall performance relative to all baselines.

\paragraph{Comparison with prompt-based methods.} Our trained decomposer improves verification performance compared to prompt-based approaches. Relative to FActScore, which produces an average of \(22.92\) subclaims, our method achieves \up{\(+6.24\)}pp F1 (\(71.75\%\) vs. \(65.51\%\)) while generating substantially fewer subclaims (\(8.14\) vs. \(22.92\)). This pattern suggests that over-decomposition may introduce noise, consistent with observations in prior work~\cite{hu-etal-2025-decomposition}. Compared to VeriScore (\(69.76\%\) F1, \(8.33\) subclaims), our approach achieves \up{\(+1.99\)}pp F1 with comparable granularity (\(8.14\) subclaims), indicating that our training procedure produces more effective decompositions at comparable granularity levels. The improvement increases to \up{\(+7.19\)}pp F1 on BINGCHAT, where both methods face challenges from longer responses (generating \(\sim\!\!30\) subclaims).
Appendix Table~\ref{app:comparisonexampels} provides examples of decomposition by different methods, highlighting some of the issues.

\paragraph{Comparison with RL-Based Methods.} 
Our approach outperforms DyDecomp~\cite{lu_optimizing_2025} overall (\up{+7.09} BAcc, \up{+5.84} macro-F1). 
On \textsc{BINGCHAT}, however, DyDecomp achieves higher F1 ($67.60\%$) but exhibits severe class imbalance: $93.69\%$ recall on \textsc{Supported} versus only $8.02\%$ on \textsc{Not Supported}$\,$(identifying fewer than 1 in 10 unsupported claims). 
Our decomposer maintains balanced recall ($61.06\%$ / $50.00\%$; \up{+41.98} on \textsc{Not Supported} vs.\ DyDecomp) despite lower F1 ($59.21\%$). 
DyDecomp trains a policy to decide whether to decompose at the \emph{response level} (no prior sentence segmentation), using verifier confidence as reward. 
This leads to minimal decompositions ($1.66$ subclaims on average), which appears to favour the \textsc{Supported} class: responses containing predominantly supported claims are easier to verify even with underdecomposed subclaims. 
However, \textsc{Not Supported} responses typically contain mixtures of supported and unsupported claims \cite{wei_long-form_2024}. Without fine-grained decomposition that isolates specific propositions, retrieval is less likely to surface disconfirming evidence for \textsc{Not Supported} cases. Our fixed sentence segmentation maintains consistent granularity, enabling balanced discrimination across both classes, which is critical for optimal verification.

\paragraph{Effect of model scale.}
Our fine-tuned 8B model achieves \(71.75\%\) overall F1, comparable to prompted larger models: Llama-3.3-70B (\(71.30\%\)) and Llama-3.1-405B (\(71.07\%\)). Notably, scaling from 70B to 405B parameters yields only \(+0.23\)pp F1 improvement when using the same prompt, while task-specific training of the 8B model provides \up{+2.58}pp improvement over its prompt-only variant (\(71.75\%\) vs. \(69.17\%\)). This pattern suggests diminishing returns to scale for prompt-only decomposition, or low sensitivity of the verifier pipeline to finer subclaim granularity. Either way, specialised training enables a smaller model to reach competitive aggregate performance.

\begin{table}[t]
\centering
\resizebox{0.85\columnwidth}{!}{%
\begin{tabular}{llrr}
\toprule
\textbf{Dataset} & \textbf{Metric} &
\multicolumn{1}{c}{\shortstack{\textbf{DAD}\\\textbf{ClearCheck}}} &
\multicolumn{1}{c}{\shortstack{\textbf{Llama-8B}\\\textbf{ClearCheck}}} \\
\midrule
\multirow{2}{*}{PerplexityAI Atom.-1} 
    & BAcc   & \textbf{72.50} & 69.50  \\
    & MacroF1 & \textbf{67.00} & 66.50  \\
\midrule
\multirow{2}{*}{PerplexityAI Atom.-2} 
    & BAcc   & \textbf{71.70} & 59.70  \\
    & MacroF1 & \textbf{70.80} & 59.80  \\
\midrule
\multirow{2}{*}{ChatGPT Atom.-1} 
    & BAcc   & \textbf{79.60} & 66.60  \\
    & MacroF1 & \textbf{79.90} & 65.40  \\
\midrule
\multirow{2}{*}{ChatGPT Atom.-2} 
    & BAcc   & 65.90 & \textbf{69.70}  \\
    & MacroF1 & \textbf{63.30} & 57.50  \\
\midrule
\multirow{2}{*}{FELM} 
    & BAcc   & \textbf{56.95} & 53.84  \\
    & MacroF1 & \textbf{62.50} & 61.75  \\
\midrule
\multirow{2}{*}{BINGCHAT} 
    & BAcc   & 51.54 & \textbf{53.17}  \\
    & MacroF1 & 18.70 & \textbf{24.90}  \\
\bottomrule
\end{tabular}
}
\caption{DAD vs. Llama-8B decomposers paired with a non-aligned ClearCheck-8B verifier. \textbf{DAD consistently outperforms Llama-8B, demonstrating that the decomposer provides value even without verifier-specific alignment}. Bold indicates the higher score per dataset and metric.}
\label{tab:clearcheck_extended}
\end{table}
\paragraph{Experiment with a non-aligned verifier.}
Our main results optimize the DAD decomposer for a specific verifier (Bespoke-MiniCheck-7B). To test whether DAD’s decompositions generalize beyond the aligned verifier, we pair it with ClearCheck-8B~\cite{seo2025verifying} and compare against a baseline using Llama-8B with the same prompt. As shown in Table~\ref{tab:clearcheck_extended}, DAD-ClearCheck outperforms Llama-8B-ClearCheck on most datasets, confirming that DAD provides value even without verifier-specific alignment. Performance remains below the aligned DAD-MiniCheck combination, indicating that verifier-specific optimization still yields the best results. The MacroF1 drop on BingChat arises from long claims (averaging 30+ subclaims) and ClearCheck’s high-confidence outputs, which tend to over-predict \textsc{Not Supported}. Performance may also be limited by using a study-wide prompt rather than a verifier-specific prompt.

\input{sections/ablation_study}

%% file: sections/ablation_study.tex
\section{Ablation Study}
 To evaluate each component’s impact on verification macro-F1, we perform an ablation study on PerplexityAI (\textit{Atomicity-1}) and FELM datasets (Table~\ref{tab:ablation}). 
 Our complete training approach—combining SFT warm-up and multi-objective rewards—yields the best overall verification performance. We highlight two key insights:

\begin{table}[t]
\centering
\resizebox{\columnwidth}{!}{%
\begin{tabular}{l*{2}{c}*{2}{c}}
\toprule
\multirow{2}{*}{\textbf{Training Configuration}} & \multicolumn{2}{c}{\textbf{PerplexityAI}} & \multicolumn{2}{c}{\textbf{FELM}} \\
\cmidrule(l){2-3}\cmidrule(r){4-5}
& \textbf{F1} & \textbf{SC} & \textbf{F1} & \textbf{SC} \\
\midrule
\quad Prompt only (no training)                  & 67.46 & 2.00 & \textbf{68.57} & 5.05 \\
\midrule
\textit{RL only (no SFT)} \\
\quad + Format + Verifier Accuracy (sparse)      & 70.86 & 1.95 & 58.94 & 3.78 \\
\quad + Format + Verifier Brier (dense)          & 72.71 & 1.49 & 64.94 & 3.40 \\
\midrule
\quad SFT only                                   & 71.55 & 2.28 & 66.39 & 5.01 \\
\midrule
\textit{SFT + RL (partial)} \\
\quad + Format + Verifier Brier                  & 71.70 & 2.03 & 66.10 & 4.43 \\
\quad + Format + Checklist                       & 69.90 & 2.28 & 67.50 & 4.97 \\
\midrule
\textit{Full SFT + RL framework (ours)} \\
\quad + Format + Verifier Brier + Checklist      & \textbf{73.15} & 2.36 & 67.23 & 5.22 \\
\bottomrule
\end{tabular}
}
\caption{Ablation study on different training setups. We report macro-F1 and Average Subclaim Count (Avg.\ SC) on PerplexityAI and FELM , plus overall F1. Best F1 per column is \textbf{bold}.}
\label{tab:ablation}
\end{table}

\begin{figure}[t]
  \centering
  \includegraphics[width=1\columnwidth]{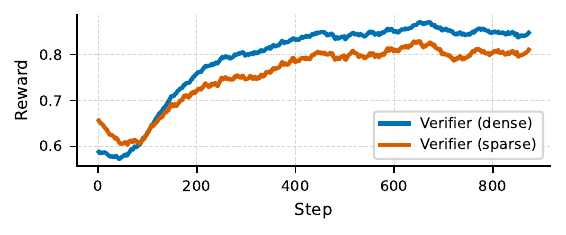}
  \caption{\textbf{Verifier training rewards (dense vs.\ sparse).}
Training the base policy with the verifier Brier reward
(\textcolor{oiBlue}{\large$\bullet$}\, dense) yields smoother, more sample-efficient learning
than the sparse accuracy reward (\textcolor{oiVermillion}{\large$\bullet$}\, sparse).}
  \label{fig:verifier-ema}
\end{figure}

\noindent\textbf{Dense Brier rewards yield more sample-efficient and stable learning.} Figure~\ref{fig:verifier-ema} shows that the base decomposer policy trained with the dense Brier reward achieves higher reward levels with fewer training steps compared to the sparse accuracy reward, while also exhibiting more stable learning dynamics and higher reward at plateau. The Brier objective leverages verifier confidence, penalizing the policy when decompositions lead to low-confidence predictions, whereas the sparse reward treats all instances equally. This richer shaping signal improves sample efficiency and translates into stronger downstream verification performance, with the dense-reward policy outperforming its sparse-reward counterpart on both PerplexityAI (\(72.71\%\) vs. \(70.86\%\)) and FELM (\(64.94\%\) vs. \(58.94\%\); Table~\ref{tab:ablation}). 

\noindent\textbf{Verifier-only reward optimisation biases the policy toward coarser decompositions, reducing generalisation.}  
Table~\ref{tab:ablation} shows that policies optimised solely with verifier rewards produce substantially fewer subclaims, indicating a tendency toward coarse decompositions. While this aligns the policy more closely with verifier preferences on the training distribution, it reduces performance on out-of-domain data: both verifier-only setups underperform on FELM relative to the base model (e.g., \(64.94\%\) vs.\ \(68.57\%\) for dense; \(58.94\%\) vs.\ \(68.57\%\) for sparse). These results suggest that relying exclusively on verifier feedback biases the policy toward in-domain verifier behaviour at the expense of generalisation. Although verifier feedback is valuable, optimising on it alone can be detrimental, and the best cross-dataset performance arises when it is combined with SFT and checklist objectives to regularise decomposition granularity. 
We provide further verification error analysis and examples in Appendix \ref{app:error_analysis}.

Additionally, we qualitatively analyzed our reasoning prompt's contribution, with example traces provided in Appendix~\ref{app:reasoning_traces}. Each decomposition step meaningfully contributes to accurate claim extraction, with intermediate reasoning (e.g., filtering non-verifiable statements, resolving pronouns, identifying relationships) systematically improving final outputs.

%% file: sections/quality_estimation.tex
\begin{figure*}[t]
    \centering
    \includegraphics[width=\linewidth]{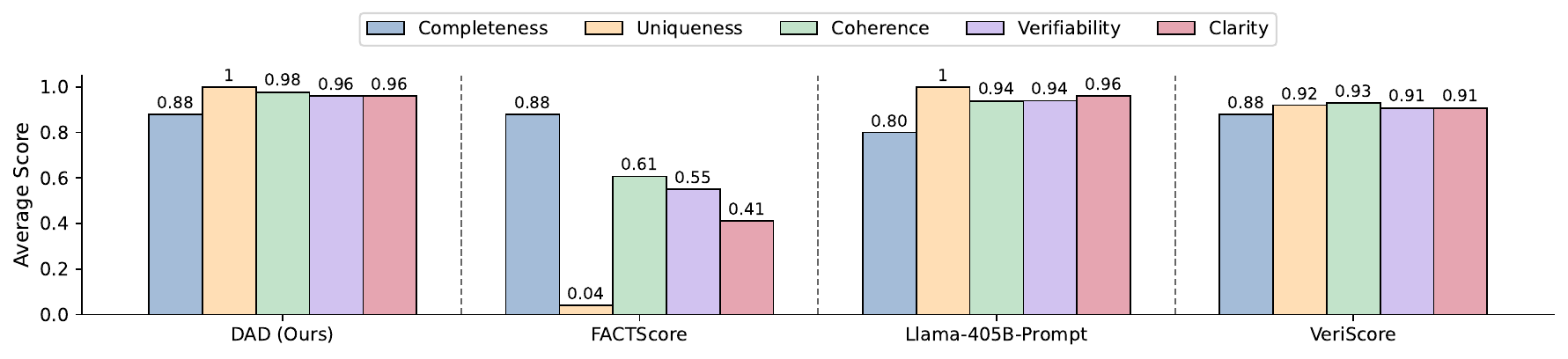}
    \caption{{\bf Human-evaluated subclaim decomposition quality across the five desiderata.} Our method achieves comparable completeness to baselines while maintaining high scores across other desiderata. Scores represent sentence-level averages of subclaim quality with higher values indicating better quality for a given desideratum.}
    \label{fig:barplots:decomposition-quality}
\end{figure*}
While recent \textit{decompose-then-verify} frameworks rely on generated subclaims from LLMs, the quality of these decompositions is not guaranteed. To assess the quality of decompositions across the desiderata for decomposition quality (see Section~\ref{sec:problem_formulation}), we manually annotated a subset of claims from our datasets, evaluating decompositions generated by our methodology and compared approaches (annotation guidelines in Appendix \ref{sec:appendixA}).

Specifically, we sampled 25 model responses across datasets, which are decomposed using our method DAD, FActScore, LLama-405B and Veriscore. These yield \(429\) sentences in total for annotation at the subclaim-level. We conduct expert annotations across the 5 desiderata: verifiability, coherence, clarity, completeness and uniqueness. Three annotators (co-authors) with expertise in natural language processing independently labeled each subclaim (or subclaim set) with binary judgement (1/0). To prevent bias, annotators were blinded to the decomposition method. Annotation agreement across the five desiderata was generally strong, with three-way agreement rates ranging from 74\% to 93\% and Fleiss' Kappa values between 0.5 and 0.88.\footnote{Detailed inter-annotator agreement results are shown in Appendix Table \ref{tab:agreement}.} We apply majority vote to determine the final labels. For any given model and desideratum (except completeness and uniqueness), scores are averaged across subclaims of the sentence. Completeness and uniqueness are assigned a single binary score at the sentence-level. See annotated examples in Appendix Table \ref{tab:bio_examples}.

\noindent\textbf{Discussion} Figure \ref{fig:barplots:decomposition-quality} presents the results. Our method which uses Llama-8B achieves a high average score for the different desiderata, demonstrating that our generated subclaim decompositions are well-formed. Notably, being significantly smaller, our method outperforms Llama-405B-Prompt across most dimensions. Interestingly, FactScore achieves high completeness, matching our method and VeriScore, but exhibits near-zero uniqueness. This suggests it generates overlapping subclaims to ensure coverage. While this might maximize information coverage, it introduces numerous trivially obvious subclaims (e.g., ``12 Feb. 2025 is a date'') or may contain information that is not directly relevant to the original claim. Incoherent subclaims provide little verification value, while redundant subclaims cause increased costs via retrieval and verifier calls. 

%% file: sections/related_works.tex
Fact verification of long-form \emph{LLM responses} commonly follows a \emph{decompose--then--verify} pipeline, in which model outputs are split into subclaims that are checked against evidence \citep{min_factscore_2023, wei_long-form_2024}. Existing decomposition approaches target different objectives; atomicity \citep{min_factscore_2023, wei_long-form_2024}, verifiability \citep{song_veriscore_2024}, decontextualisation \citep{gunjal_molecular_2024}, and efficiency \citep{rajendhran2025verifastscorespeedinglongformfactuality}—but are primarily designed for benchmarking LLM factuality rather than accurate verification. This yields two key limitations: \textbf{verifier misalignment} (decomposition objectives are not optimised with respect to the verifier, which is treated as fixed; \citealp{lu_optimizing_2025}) and \textbf{decomposition quality} (decomposition can introduce noise that degrades downstream verifier accuracy; \citealp{hu-etal-2025-decomposition}).

Recent work addresses these limitations separately. To improve decomposition quality, \citet{wanner2024dndscoredecontextualizationdecompositionfactuality, gunjal_molecular_2024} focus on generating \emph{molecular facts}—i.e., minimal, fully decontextualised subclaims; \citet{liu2025verifactenhancinglongformfactuality} aim to capture missing or incomplete facts; and \citet{metropolitansky_towards_2025} propose a multi-stage pipeline that separates selection, disambiguation, and extraction. While these methods improve different decomposition \emph{desiderata}, they do not address verifier alignment and often require complex pipelines with closed-source models. Conversely, \citet{lu_optimizing_2025} tackle verifier misalignment by training, via PPO, a separate policy that decides whether to decompose based on verifier confidence as reward. However, this approach learns only \emph{when} to decompose while leaving the decomposer unchanged, achieving alignment without improving decomposition quality. \textbf{Our approach addresses both limitations jointly}. We reframe decomposition as a sequential reasoning task in which the decomposer performs explicit intermediate reasoning steps before generating subclaims within a single model call. To achieve verifier alignment \emph{and} high-quality decomposition, we propose a \textbf{multi-objective reward} that combines a verifier-based signal with an explicit decomposition-quality checklist, and we train the decomposer policy with GRPO \citep{shao2024deepseekmathpushinglimitsmathematical}. This joint optimisation yields consistent gains in verification accuracy and in human-evaluated decomposition quality.
Appendix~\ref{app:literature_table} provides a comparative table of our method versus prior work, emphasizing novelty.

%% file: sections/conclusion.tex
We address the challenge of complex claim verification by jointly optimizing decomposition for quality and verifier alignment. Our novel multi-objective reward, combined with structured sequential reasoning, enables efficient generation of high-quality subclaims and improves verification accuracy. Our trained decomposer improves verification accuracy while maintaining balanced subclaim granularity across diverse evaluation settings.

%% file: sections/limitations.tex
While our approach demonstrates improvements in claim decomposition and verification, several limitations remain. First, we only test our approach on a fixed verifier and are limited by the available reference datasets, which impacts the extent to which we can improve and generalize verification accuracy. Future work should assess the framework across a broader range of verifiers and knowledge sources. Second, supervised pre-training relied on synthetic exemplars from a single teacher model, potentially introducing bias and limiting decomposition diversity. Third, our multi-objective reward and LLM-as-a-Judge evaluations, while scalable, may not fully capture nuanced aspects of decomposition quality; human evaluation was limited in scope. Fourth, our LLM-as-a-Judge element may introduce some undesired bias in the training process, which could be assessed by further experimentation with alternative large models as judges. Fifth, the approach was tested primarily on English-language data and Wikipedia/Google Search, leaving multilingual and domain-specific verification as future work. Sixth, our evaluation did not include multi-hop claim verification benchmarks, which are important for assessing performance on complex reasoning tasks involving multiple interconnected facts. Lastly, sentence-level segmentation may not optimally capture all factual relationships, especially those spanning multiple sentences or extend across the entire document. Investigating more adaptive segmentation and decomposition strategies could further improve performance on complex long-form texts.

%% file: sections/appendixA.tex
\subsection{Overview}

This framework evaluates decomposed subclaims on five qualitative dimensions. Each dimension is scored in a binary fashion (1/0) per annotator. Scores are averaged across dimensions to yield a subclaim-level score, and across subclaims to yield model-level performance.

\subsection{Dimensions}

\paragraph{Verifiability (Specificity)}
Can the subclaim be checked against a reliable reference (e.g., Google Search)?
\begin{itemize}[noitemsep, topsep=4pt]
    \item \textbf{Score 1}: Verifiable.
    \item \textbf{Score 0}: Vague or unverifiable.
\end{itemize}

\textbf{Examples:}
\begin{itemize}[noitemsep, topsep=4pt]
    \item ``Tesla’s Q4 2024 revenue was \$25 billion.'' (\textbf{Verifiable})
    \item ``Tesla is very successful.'' (\textbf{Not verifiable})
\end{itemize}

\vspace{1em}

\paragraph{Faithfulness (Coherence / Entailment)}
Does the subclaim preserve the meaning of the original sentence without adding, removing, or distorting information? If the original sentence encodes a relationship (causal, temporal, comparative, conditional, attribution), the decomposition must preserve that relationship.

\begin{itemize}[noitemsep, topsep=4pt]
    \item \textbf{Score 1}: Fully faithful to the source.
    \item \textbf{Score 0}: Alters or misrepresents the source or does not preserve relationship.
\end{itemize}

\textbf{Examples:}
\begin{itemize}[noitemsep, topsep=4pt]
    \item \textbf{Original claim}: The Fed raised interest rates by 0.25\% in March.
    \begin{itemize}
        \item \textbf{Subclaim}: The Fed raised interest rates in March. (\textbf{Faithful simplification})
        \item \textbf{Subclaim}: The Fed raised interest rates by 0.5\% in March. (\textbf{Distorted detail})
    \end{itemize}
    \item \textbf{Original claim}: James was born in 1998.
    \begin{itemize}
        \item \textbf{Bad Decomposition}: James was born. 1998 exists.
        \item \textbf{Good Decomposition}: James was born in 1998. (Temporal relationship preserved)
    \end{itemize}
\end{itemize}

\vspace{1em}

\paragraph{Clarity (Self-sufficiency / Decontextualization + Non-triviality)}
Is the subclaim clear, unambiguous, and understandable without additional context?

\begin{itemize}[noitemsep, topsep=4pt]
    \item \textbf{Score 1}: Self-contained and precise.
    \item \textbf{Score 0}: Ambiguous (e.g., pronouns, vague references) or trivial (e.g., ``X exists'').
\end{itemize}

\textbf{Examples:}
\begin{itemize}[noitemsep, topsep=4pt]
    \item ``Microsoft acquired Activision Blizzard in 2022.'' (\textbf{Clear})
    \item ``They acquired the company in 2022.'' (\textbf{Unclear reference})
    \item ``James is born.'' (\textbf{Trivial})
    \item ``1998 exists.'' (\textbf{Trivial})
\end{itemize}

\vspace{1em}

\paragraph{Coverage (Completeness)}
Does the set of subclaims for a sentence capture all verifiable facts expressed in the original sentence?

\begin{itemize}[noitemsep, topsep=4pt]
    \item \textbf{Score 1}: All facts are captured.
    \item \textbf{Score 0}: One or more facts are missing.
\end{itemize}

\textbf{Examples:}
\begin{itemize}[noitemsep, topsep=4pt]
    \item \textbf{Original}: California and New York implemented plastic bag bans.
    \item \textbf{Subclaims} (Coverage = 1):
    \begin{itemize}[noitemsep]
        \item California implemented a plastic bag ban.
        \item New York implemented a plastic bag ban.
    \end{itemize}
    \item \textbf{Subclaims} (Coverage = 0; New York fact missing):
    \begin{itemize}
        \item California implemented a plastic bag ban.
    \end{itemize}
\end{itemize}

\vspace{1em}

\paragraph{Uniqueness}
Do the extracted subclaims avoid duplicating the same fact in different word orders or paraphrases?

\begin{itemize}[noitemsep, topsep=4pt]
    \item \textbf{Score 1}: Each subclaim adds unique factual content, with no overlap in meaning.
    \item \textbf{Score 0}: One or more subclaims repeat the same fact (e.g., ``A and B'' vs. ``B and A'').
\end{itemize}

\textbf{Examples:}
\begin{itemize}[noitemsep, topsep=4pt]
    \item \textbf{Original}: California and New York implemented a plastic bag ban.
    \item \textbf{Subclaims} (Unique facts; Uniqueness score = 1):
    \begin{itemize}
        \item California implemented a plastic bag ban.
        \item New York implemented a plastic bag ban.
    \end{itemize}
    \item \textbf{Original}: Microsoft and Google released AI chatbots.
    \item \textbf{Subclaims} (Duplication; Uniqueness score = 0):
    \begin{itemize}
        \item Microsoft released an AI chatbot.
        \item Google released an AI chatbot.
        \item AI chatbots were released by Microsoft and Google.
    \end{itemize}
\end{itemize}

\vspace{1em}

\subsection{Scoring Methodology}

Each subclaim receives a score for Verifiability, Faithfulness, and Clarity.  
All subclaims together for each original sentence receive a Coverage and Uniqueness score.

\section{Dataset Details}
\label{appendix:dataset_details}
\begin{table}[h!]
\centering
\setlength{\tabcolsep}{4pt}
\resizebox{\linewidth}{!}{%
\begin{tabular}{llcrl}
\toprule
\textbf{Granularity} & \textbf{Dataset} & \textbf{Atomicity} & \textbf{Size} & \textbf{KS} \\
\midrule
\multirow{4}{*}{Sentence} 
& PerplexityAI & 1 & 300 & Wiki \\
& ChatGPT      & 1 & 249 & Wiki \\
& PerplexityAI & 2 & 149 & Wiki \\
& ChatGPT      & 2 & 115 & Wiki \\
\cmidrule(lr){1-5}
\multirow{2}{*}{Response} 
& FELM         & --- & 184 & Google \\
& BINGCHAT     & --- & 396 & Google \\
\bottomrule
\end{tabular}%
}
\caption{Evaluation datasets by label granularity, atomicity, and size. Abbreviations: KS = Knowledge Source; Wiki = Wikipedia.}
\label{tab:evaluation-datasets}
\end{table}

\begin{table*}[ht]
\centering
\begin{tabular}{lccccc}
\hline
\textbf{Desiderata} & \textbf{3-way (\%)} & \textbf{Fleiss' $\kappa$ } & \textbf{Pairwise (\%)} & \textbf{$\kappa$} & \textbf{r} \\
\hline
\textbf{Completeness} & 80.8 & 0.5 & A1-A3: 85.0 & 0.5 & 0.5 \\
                      &      &     & A1-A2: 85.9 & 0.5 & 0.5 \\
                      &      &     & A3-A2: 90.9 & 0.6 & 0.6 \\
\textbf{Uniqueness}   & 92.9 & 0.9 & A1-A3: 94.0 & 0.8 & 0.8 \\
                      &      &     & A1-A2: 93.9 & 0.8 & 0.8 \\
                      &      &     & A3-A2: 97.0 & 0.9 & 0.9 \\
\textbf{Verifiability}& 81.9 & 0.7 & A1-A3: 90.9 & 0.8 & 0.8 \\
                      &      &     & A1-A2: 86.3 & 0.6 & 0.6 \\
                      &      &     & A3-A2: 86.3 & 0.6 & 0.6 \\
\textbf{Coherence}    & 73.6 & 0.5 & A1-A3: 81.1 & 0.5 & 0.5 \\
                      &      &     & A1-A2: 85.3 & 0.6 & 0.6 \\
                      &      &     & A3-A2: 80.9 & 0.5 & 0.5 \\
\textbf{Clarity} & 81.2 & 0.7 & A1-A3: 88.3 & 0.7 & 0.7 \\
                      &      &     & A1-A2: 85.6 & 0.7 & 0.7 \\
                      &      &     & A3-A2: 88.0 & 0.7 & 0.7 \\
\hline
\end{tabular}
\caption{Inter-annotator agreement statistics across the five desiderata. 3-way agreement shows percentage agreement among all three annotators (A1, A2, A3) with Fleiss' $\kappa$ for multi-annotator reliability. Pairwise statistics include percentage agreement, Cohen's $\kappa$, and Pearson correlation coefficient (r) for each annotator pair.}
\label{tab:agreement}
\end{table*}

\section{Annotation Examples}
Table~\ref{tab:bio_examples} shows annotated examples of decompositions produced by DAD. 
\begin{table*}[t]
\centering
\begin{tabular}{p{4cm} p{12cm}}
\hline
\textbf{Question} & Businessman, a Jewish American, grew up and lived in New York City, associated with the Democratic Party, and eventually became very famous. His name is "Bernie"-what? \\
\hline
\textbf{Sentence} & Bernie Sanders. \\
\hline
\textbf{Subclaims} & Bernie Sanders \\
\hline
\textbf{Human Annotation Scores} & 
\begin{tabular}{lccccc}
\textbf{Completeness} & \textbf{Redundancy} & \textbf{Verifiability} & \textbf{Coherence} & \textbf{Clarity} \\
Majority & Majority & Average & Average & Average \\
0 & 1 & 0 & 1 & 0 \\
\end{tabular} \\
\hline
\multicolumn{2}{c}{\rule{0pt}{12pt}} \\
\hline
\textbf{Question} & Tell me a bio of Paul Kruger. \\
\hline
\textbf{Sentence} & Kruger was a prominent figure in the Boer War (1899-1902), leading the Boers in their defense against the British Empire. Kruger was a staunch believer in Boer independence and was instrumental in securing the Transvaal's independence from British rule in 1881. \\
\hline
\textbf{Subclaims} & 
1. Paul Kruger was a prominent figure in the Boer War (1899-1902). \newline
2. Paul Kruger led the Boers in their defense against the British Empire during the Boer War. \newline
3. Paul Kruger was a staunch believer in Boer independence. \newline
4. Paul Kruger was instrumental in securing the Transvaal's independence from British rule in 1881. \\
\hline
\textbf{Human Annotation Scores} & 
\begin{tabular}{lccccc}
\textbf{Completeness} & \textbf{Redundancy} & \textbf{Verifiability} & \textbf{Coherence} & \textbf{Clarity} \\
Majority & Majority & Average & Average & Average \\
1 & 1 & 1 & 1 & 1 \\
\end{tabular} \\
\hline
\multicolumn{2}{c}{\rule{0pt}{12pt}} \\
\hline
\textbf{Question} & Tell me a bio of Rakhi Sawant. \\
\hline
\textbf{Sentence} & Rakhi Sawant established herself as a sex symbol in Bollywood. She was a contestant on the first season of the Indian reality television series Bigg Boss 1 (2006) and a challenger and finalist in Bigg Boss 14 (2020). \\
\hline
\textbf{Subclaims} & 
1. Rakhi Sawant established herself as a sex symbol in the Indian film industry. \newline
2. Rakhi Sawant was a contestant on the first season of the Indian reality television series Bigg Boss 1 (2006). \newline
3. Rakhi Sawant was a challenger and finalist in Bigg Boss 14 (2020). \\
\hline
\textbf{Human Annotation Scores} & 
\begin{tabular}{lccccc}
\textbf{Completeness} & \textbf{Redundancy} & \textbf{Verifiability} & \textbf{Coherence} & \textbf{Clarity} \\
Majority & Majority & Average & Average & Average \\
0 & 1 & 1 & 0.67 & 1 \\
\end{tabular} \\
\hline
\end{tabular}
\caption{Subclaims and Human Annotation Scores for Example Questions}
\label{tab:bio_examples}
\end{table*}

\section{Comparative Analysis with Existing Literature}
\label{app:literature_table}
Table \ref{app:comp_analysis} provides a side-by-side comparison of our method against relevant prior works, summarizing key differences and improvements. 
\begin{table*}[ht]
\centering
\begin{tabular}{p{4cm} p{5cm} p{7cm}}
\toprule
\textbf{Method} 
& \textbf{What is Similar to Our Approach} 
& \textbf{What is Different / Missing Compared to Ours} \\
\midrule
\textbf{FActScore} \cite{min_factscore_2023} 
& Focuses on atomic fact extraction; uses selection and extraction steps. 
& Over-decomposes (too many subclaims); does not optimize for verifier alignment; lacks explicit decontextualization and relationship identification. \\

\textbf{VeriScore} \cite{song_veriscore_2024}
& Emphasizes verifiability in claim extraction; uses prompt-based decomposition. 
& Does not explicitly decontextualize claims; lacks relationship identification; does not jointly optimize decomposition and verifier alignment. \\

\textbf{Molecular Facts} \cite{gunjal_molecular_2024}
& Targets fully decontextualized, minimal subclaims. 
& Focuses only on decontextualization; does not address selection, relationship identification, or verifier alignment. \\

\textbf{Claimify} \cite{metropolitansky_towards_2025} 
& Separates selection, disambiguation, and extraction into distinct steps. 
& Uses complex, multi-stage process with closed-source models; does not jointly optimize decomposition and verifier alignment; lacks efficiency of single model call. \\

\textbf{DyDecomp} \cite{lu_optimizing_2025} 
& RL-based optimization for when to decompose, using verifier feedback. 
& Only decides when to decompose, not how; does not improve decomposition quality or address relationships and decontextualization. \\

\textbf{DndScore} \cite{wanner-etal-2024-closer} 
& Focuses on decontextualization and decomposition for factuality verification. 
& Does not integrate selection, relationship identification, and verifier alignment in a unified framework. \\

\textbf{VeriFastScore} \cite{rajendhran_verifastscore_2025} 
& Aims for efficient long-form factuality evaluation with refined fact extraction. 
& Prioritizes efficiency; does not address relationship identification or joint optimization with verifier. \\
\bottomrule
\end{tabular}
\caption{Comparison of Decomposer-Focused Methods: Similarities and Differences with Our Approach}
\label{app:comp_analysis}
\end{table*}

\section{Error Analysis}
\label{app:error_analysis}
\input{sections/error_analysis}
\begin{table*}[t]
    \centering
    \renewcommand{\arraystretch}{1.2}
    \begin{tabularx}{\textwidth}{l X}
        \toprule
        \textbf{Field} & \textbf{Content} \\
        \midrule
        \multicolumn{2}{l}{\textbf{Example 1: Frederick Howard Taylor}} \\
        \midrule
        Question & Tell me a bio of Frederick Howard Taylor. \\
        Original Claim & Howard Taylor attended the Royal College of Surgeons where he received three high honors in his postgraduate studies. \\
        Subclaims & 
        1. Howard Taylor attended the Royal College of Surgeons where he received three high honors in his postgraduate studies. \\
        Ground Truth Label & Not Supported (NS) \\
        Model Prediction & Supported (S) \\
        Evidence & 
        Taylor was a Fellow of the Royal College of Surgeons and received three high honors. \\
        Error Cause & 
        Decomposer could have decomposed the claim further. Verifier conflated "fellowship" with "attendance," supporting the claim without direct evidence of attendance. This reflects a reasoning failure and insufficiently strict evidence matching. \\
        \midrule
        \multicolumn{2}{l}{\textbf{Example 2: Mauro Icardi}} \\
        \midrule
        Question & Tell me a bio of Mauro Icardi. \\
        Original Claim & He plays as a striker for Süper Lig club Galatasaray, on loan from Ligue 1 club Paris Saint-Germain and the Argentina national team. \\
        Subclaims & 
        1. Mauro Icardi plays as a striker for Galatasaray. \newline
        2. Mauro Icardi is on loan from Paris Saint-Germain. \newline
        3. Mauro Icardi is on loan from the Argentina national team. \\
        Ground Truth Label & Not Supported (NS) \\
        Model Prediction & Supported (S) \\
        Evidence & 
        The evidence confirms Icardi's club status and loan from PSG, but there is no such thing as a loan from a national team in football. \\
        Error Cause & 
        Lack of world knowledge in the verifier led to its failure to recognize the impossibility of a national team loan. \\
        \midrule
        \multicolumn{2}{l}{\textbf{Example 3: Eric Hacker}} \\
        \midrule
        Question & Tell me a bio of Eric Hacker. \\
        Original Claim & As of March 2023, Eric Hacker is a free agent. \\
        Subclaims & 
        1. Eric Hacker is a free agent as of March 2023. \\
        Ground Truth Label & Not Supported (NS) \\
        Model Prediction & Supported (S) \\
        Evidence & 
        Eric Lynn Hacker (born March 26, 1983) is an American professional baseball pitcher who is a free agent. Several mentions of him being a free agent in earlier periods. However, no mention of March 2023. \\
        Error Cause & 
        Verifier overgeneralized or relied on outdated information, supporting the claim without up-to-date evidence. \\
        \bottomrule
    \end{tabularx}
     \caption{Detailed Verification Error Analysis with Subclaims and Evidence}
      \label{tab:error_analysis_blocks}
\end{table*}

\section{Reasoning Traces Analysis}
\label{app:reasoning_traces}
Following the error analysis, we qualitatively examine how the sequential decomposition prompt influences intermediate reasoning and final outputs, focusing on the four-step process described in Section~\ref{sec:methods}: Step~1---Claim detection ($S_i \rightarrow$ \textit{VerifiableSentence}); Step~2---Decontextualization (\textit{VerifiableSentence} $\rightarrow$ \textit{DecontextualizedSentence}); Step~3---Relationship identification (\textit{DecontextualizedSentence} $\rightarrow$ \textit{Relationships}); and Step~4---Claim extraction (\textit{DecontextualizedSentence} + \textit{Relationships} $\rightarrow$ \textit{Claims}). Appendix Table~\ref{tab:error_analysis_thinking_blocks} presents representative cases illustrating the contribution of each step.

In Example~1, Step~1 correctly excludes a lack-of-information statement (``but little else is known about her current whereabouts or status'') and retains only the verifiable proposition. In Example~2, Step~2 confirms that all pronouns (e.g., ``he'') are resolved to explicit entity names, yielding a self-contained \textit{DecontextualizedSentence}; in this instance, the target entity was already explicit in the \textit{VerifiableSentence} produced by Step~1. Step~3 is effective in both examples: in Example~1, it identifies temporal and expansion relations; in Example~2, it identifies temporal and membership/part-of relations. These intermediate annotations are crucial for Step~4, which splits the sentence into minimal, fully decontextualized factual units while preserving identified relationships. In both cases, the sentence-level label assigned by the verifier matches the ground-truth annotation.

\begin{table*}[t]
    \centering
    \renewcommand{\arraystretch}{1.2}
    \begin{tabularx}{\textwidth}{l X}
        \toprule
        \textbf{Field} & \textbf{Content} \\
\midrule
        \multicolumn{2}{l}{\textbf{Example 1: Lina Medina [Dataset: ChatGPT,  Atomicity 1]}} \\
        \midrule
        Question & Tell me a bio of Lina Medina. \\
        Original Claim & Lina lived a relatively private life after the birth of her son and later married in 1970, but little else is known about her current whereabouts or status. \\
	Reasoning & \small STEP 1: The target sentence contains verifiable claims about Lina Medina's life after the birth of her son, specifically that she lived a relatively private life and later married in 1970. \hl{However, the sentence also mentions that little else is known about her current whereabouts or status, which is a lack-of-information statement and should be excluded}. The verifiable sentence is: "Lina lived a relatively private life after the birth of her son and later married in 1970."

STEP 2: To decontextualize the sentence, we need to ensure it is clear, self-contained, and contextually relevant. The sentence already mentions Lina's name, and the context provides the necessary information about her son's birth. The decontextualized sentence is: "Lina Medina lived a relatively private life after the birth of her son and later married in 1970."

STEP 3: Analyzing the decontextualized sentence, we can identify the following relationships: \hl{
- Temporal relationship: Lina lived a private life after the birth of her son and later married in 1970 (time sequence). 
- Expansion relationship: The sentence provides additional information about Lina's life after the birth of her son (elaboration).}

STEP 4: To split the claim into the smallest possible factual units, we can break it down into two separate claims: - Claim 1: Lina Medina lived a relatively private life after the birth of her son. - Claim 2: Lina Medina married in 1970. \\ 
	Subclaims & 1. Lina Medina lived a relatively private life after the birth of her son. 

2. Lina Medina married in 1970. \\
	Model prediction & Not Supported (NS) \\ 
	Ground truth & Not Supported (NS) \\ 

        \midrule
        \multicolumn{2}{l}{\textbf{Example 2: Ronaldo (Brazilian footballer) [Dataset: PerplexityAI, Atomicity 1]}} \\
        \midrule
        Question & Tell me a bio of Ronaldo (Brazilian footballer). \\
        Original Claim & At age 17, he was the youngest member of the Brazilian squad that won the 1994 FIFA World Cup. \\
	Reasoning & \small STEP 1: The target sentence contains a specific and verifiable proposition about Ronaldo's age and his participation in the 1994 FIFA World Cup. This meets the criteria for a verifiable claim. VerifiableSentence: At age 17, Ronaldo was the youngest member of the Brazilian squad that won the 1994 FIFA World Cup.

STEP 2: To decontextualize the sentence, we need to ensure it stands alone as a query-relevant, unambiguous statement. \hl{The sentence already mentions Ronaldo by name and specifies the event (1994 FIFA World Cup) and his role} (youngest member of the Brazilian squad). No additional information from the Question or Context is needed to resolve ambiguity. DecontextualizedSentence: At age 17, Ronaldo was the youngest member of the Brazilian squad that won the 1994 FIFA World Cup.

STEP 3: Identify relationships in the sentence. \hl{There is a temporal relationship indicated by "At age 17,"} which specifies Ronaldo's age at the time of the event. \hl{There is also a membership/part-of relationship}, as Ronaldo is part of the Brazilian squad. Relationships: Temporal, Membership/Part-of.

STEP 4: Split the claim into the smallest possible factual units. From the DecontextualizedSentence and identified relationships, we can split the claim into two specific, verifiable, and fully decontextualized claims. Claims: 
1. Ronaldo was 17 years old when the Brazilian squad won the 1994 FIFA World Cup.
2. Ronaldo was the youngest member of the Brazilian squad that won the 1994 FIFA World Cup. \\
        Subclaims & 
        1. Ronaldo was 17 years old when the Brazilian squad won the 1994 FIFA World Cup. 2. Ronaldo was the youngest member of the Brazilian squad that won the 1994 FIFA World Cup. \\ 
	Model prediction & Supported (S) \\ 
	Ground truth & Supported (S) \\ 
\bottomrule
    \end{tabularx}
     \caption{Decomposition analysis with reasoning traces. Highlighted segments are discussed in Appendix \ref{app:reasoning_traces}.}
      \label{tab:error_analysis_thinking_blocks}
\end{table*}

% Body (place where you want the tables)
\begin{table*}[t]
%\caption{Structured examples with vertical field layout and compacted factscore lists}
%\small
\begin{tabularx}{\textwidth}{L Y}
\toprule
\textbf{Field} & \textbf{Content} \\
\midrule
\multicolumn{2}{l}{\textbf{Example: Harrison Ford}}\\
\textbf{Question} & Tell me a bio of Harrison Ford. \\
\textbf{Text} & He was born on July 13, 1942, in Chicago, Illinois, to former radio actress Dorothy and advertising executive John William "Christopher" Ford. \\
\textbf{Label} & Supported \vspace{0.2cm} \\ \hline 
 \vspace{0.1cm}  \textbf{FactScore (FS) Subclaims} & \vspace{0.1cm} 
[ `\hl{He} was born on July 13, 1942.', `\hl{He} was born in 1942.', \hl{`1942 is a year.'}, \hl{`July 13 is a date.'}, \hl{`July 13 is a date in 1942.'}, \hl{`He} was born on July 13.', `He was born in July 1942.', \hl{`July is a month.'}, `July is a month of 1942.', `He was born in July.', `He was born in July 1942.', `He was born in 1942.', <.. repeated variants of `He was born in the year of his birth, 1942.' ..> ] \\
%subclaim\_predictions\_factscore &
%[ 1, 1, 1, 1, 1, 1, 1, 1, 0, 1, 1, 1, 1, 0, 1, 1, <.. many 0s ..> ] \\
\textbf{Prediction using FS subclaims}  & Not Supported  \\ \hline
\vspace{0.1cm} \textbf{VeriScore (VS) Subclaims} & \vspace{0.1cm}
[`Harrison Ford was born on July 13, 1942.', `Harrison Ford was born in Chicago, Illinois.', \hl{`Harrison Ford's father was a former radio actress named Dorothy.'}, `Harrison Ford's father was an advertising executive named John William "Christopher" Ford.' ] \\ 
%VeriScore (VS) Subclaims  & [ 1, 1, 0, 1 ] \\
\textbf{Prediction using VS subclaims} & Not Supported\\ \hline
\vspace{0.1cm} \textbf{DAD Subclaims} & \vspace{0.1cm}
[`Harrison Ford was born on July 13, 1942.', `Harrison Ford was born in Chicago, Illinois.', `Harrison Ford's mother was a former radio actress named Dorothy.', `Harrison Ford's father was an advertising executive named John William "Christopher" Ford.' ] \\
%subclaim\_predictions\_dad & [ 1, 1, 1, 1 ] \\
\textbf{Prediction using DAD subclaims} & Supported \\
\textbf{DAD Reasoning Trace}  &
\reasoning{<think>
STEP 1: The target sentence contains specific and verifiable information about Harrison Ford's birth, including the date, place, and parents. VerifiableSentence: He was born on July 13, 1942, in Chicago, Illinois, to former radio actress Dorothy and advertising executive John William "Christopher" Ford.

STEP 2: DecontextualizedSentence: Harrison Ford was born on July 13, 1942, in Chicago, Illinois, to former radio actress Dorothy and advertising executive John William "Christopher" Ford.

STEP 3: Relationships: Membership/Part-of (Harrison Ford's birth family).

STEP 4: Claims:
1. Harrison Ford was born on July 13, 1942.
2. Harrison Ford was born in Chicago, Illinois.
3. Harrison Ford's mother was a former radio actress named Dorothy.
4. Harrison Ford's father was an advertising executive named John William "Christopher" Ford.
</think>
<output>["Harrison Ford was born on July 13, 1942.", "Harrison Ford was born in Chicago, Illinois.", "Harrison Ford's mother was a former radio actress named Dorothy.", "Harrison Ford's father was an advertising executive named John William \"Christopher\" Ford."]</output>} \\
\bottomrule
\end{tabularx}
\caption{Examples comparing decompositions across several methods. Highlighted issues with decompositions. }
\end{table*}

\begin{table*}[t]
\ContinuedFloat
%\small
\begin{tabularx}{\textwidth}{L Y}
\toprule
\textbf{Field} & \textbf{Content} \\
\midrule
\multicolumn{2}{l}{\textbf{Example: Scientist (musician)}}\\
\textbf{Question} & Tell me a bio of Scientist (musician). \\
\textbf{Text} & He was a protégé of King Tubby and worked alongside other pioneers of the genre such as Bunny Lee, Philip Smart, Pat Kelly, and Prince Jammy. \\
\textbf{Label} & Supported \vspace{0.2cm} \\ \hline
\vspace{0.1cm} \textbf{FactScore (FS) Subclaims} & \vspace{0.1cm}
[ `\hl{He} was a protégé of King Tubby.', \hl{`King Tubby is a person.'}, `\hl{He} was a protégé.', `\hl{He} was a protégé of King Tubby.', `King Tubby is a person who was a mentor.', `He worked alongside other pioneers of the genre.', \hl{`The genre is a type of music.'}, `Bunny Lee is a pioneer of the genre.', `He worked alongside Bunny Lee.', `Philip Smart is a pioneer of the genre.', `He worked alongside Philip Smart.', `Pat Kelly is a pioneer of the genre.', `He worked alongside Pat Kelly.', `Prince Jammy is a pioneer of the genre.', `He worked alongside Prince Jammy.' ] \\
% \textbf{FS Subclaim Predictions} & [ 1, 1, 1, 1, 1, 1, 1, 0, 0, 1, 0, 1, 0, 0, 1 ] \\
\textbf{Prediction using FS subclaims}  & Not Supported \\ \hline
\vspace{0.1cm} \textbf{VeriScore (VS) Subclaims} & \vspace{0.1cm}
[ `Hopeton Overton Brown, known as Scientist, was a protégé of King Tubby.', `Hopeton Overton Brown, known as Scientist, worked alongside King Tubby.', `Hopeton Overton Brown, known as Scientist, worked alongside Bunny Lee.', `Hopeton Overton Brown, known as Scientist, worked alongside Philip Smart.', `Hopeton Overton Brown, known as Scientist, worked alongside Pat Kelly.', `Hopeton Overton Brown, known as Scientist, worked alongside Prince Jammy.' ] \\ 
% \textbf{VS Subclaim Predictions} & [ 1, 1, 1, 0, 1, 1 ] \\
\textbf{Prediction using VS subclaims} & Not Supported \\ \hline
\vspace{0.1cm} \textbf{DAD Subclaims} & \vspace{0.1cm}
[ `Hopeton Overton Brown (Scientist) was a protégé of King Tubby.', `Hopeton Overton Brown (Scientist) worked alongside other pioneers of the genre such as Bunny Lee, Philip Smart, Pat Kelly, and Prince Jammy.' ] \\
% \textbf{DAD Subclaim Predictions} & [ 1, 1 ] \\
\textbf{Prediction using DAD subclaims} & Supported \\
\textbf{DAD Reasoning Trace}  &
\reasoning{<think>
STEP 1: VerifiableSentence: He was a protégé of King Tubby and worked alongside other pioneers of the genre such as Bunny Lee, Philip Smart, Pat Kelly, and Prince Jammy.

STEP 2: DecontextualizedSentence: Hopeton Overton Brown (Scientist) was a protégé of King Tubby and worked alongside other pioneers of the genre such as Bunny Lee, Philip Smart, Pat Kelly, and Prince Jammy.

STEP 3: Relationships: Attribution; Membership/Part-of.

STEP 4: Claims:
- Hopeton Overton Brown (Scientist) was a protégé of King Tubby.
- Hopeton Overton Brown (Scientist) worked alongside other pioneers of the genre such as Bunny Lee, Philip Smart, Pat Kelly, and Prince Jammy.
</think>
<output>["Hopeton Overton Brown (Scientist) was a protégé of King Tubby.", "Hopeton Overton Brown (Scientist) worked alongside other pioneers of the genre such as Bunny Lee, Philip Smart, Pat Kelly, and Prince Jammy."]</output>} \\
\bottomrule
\end{tabularx}
\caption{Examples comparing decompositions across several methods. Highlighted issues with decompositions. (continued)}
\label{app:comparisonexampels}
\end{table*}

\section{Implementation Details}
\label{appendix:implementation_details}

\paragraph{Sentence Segmentation.}
We perform sentence segmentation using \texttt{spaCy}\footnote{\url{https://spacy.io/}} to split model responses into individual sentences. For each target sentence, we construct a local context window including the $p=2$ preceding and $f=2$ following sentences, truncated to document boundaries as needed. This segmented structure forms the input tuples $(Q, C_i, S_i)$ used for decomposition.

\paragraph{Synthetic Data Generation.}
We generate synthetic decompositions using Llama-3.1-405B-Instruct via Amazon Bedrock with temperature \(0.35\) and top-p \(0.9\). For each of the 15.7k sampled sentences, we construct a prompt containing the original question, surrounding context (preceding and following sentences), and the target sentence, then generate decompositions using our decomposition prompt shown in Appendix ~\ref{appendix:prompts}. 

\paragraph{SFT Warmup.}
We fine-tune Llama-3.1-8B-Instruct using LoRA \citep{hu2022lora} with rank \(r=32\) and \(\alpha=64\), targeting all attention and MLP projection layers. We train for 1 epoch on 13.7k examples with batch size 32 (8 per device, 4 gradient accumulation steps), learning rate \(2 \times 10^{-4}\) with cosine decay and 10\% warmup, and weight decay 0.01. We use AdamW 8-bit optimizer and mixed precision training. Training uses the Llama-3.1 chat template with standard roles and applies loss only on assistant tokens. We train on a single A100 80GB GPU for approximately 8 hours using the Unsloth implementation~\cite{unsloth}\footnote{\url{https://github.com/unslothai/unsloth}}.

\paragraph{RL Implementation.}
We fine-tune the SFT warm-up policy (trained for one epoch) using the Dr.GRPO implementation ~\citep{liu2025understanding}. Training runs for~600~steps with a per-GPU batch size of~4 and gradient accumulation of~8, yielding an effective batch of~256~prompts per update. Each prompt generates~8~completions per update group. Prompts are truncated to ~2{,}048~tokens and completions to~3{,}072~tokens within a ~4{,}096-token context window. Optimization follows a constant-with-warmup schedule (20~warmup steps) with a learning rate of~\(2\times10^{-6}\).  For memory efficiency, LoRA~\citep{hu2022lora} adapters with rank~\(r{=}16\) and \(\alpha{=}32\) are applied to all attention and MLP projection layers. The policy is implemented and trained using the Verifiers implementation~\cite{brown_verifiers_2025} \footnote{\url{https://github.com/PrimeIntellect-ai/verifiers}}. Training is conducted on an 8$\times$A100 (40GB) setup and completes in approximately 15~hours.

\section{Reward Implementation Details}
\label{appendix:reward-implementation}
We provide the implementation details of our reward terms in this section. 

\subsection{Checklist Reward}
\begin{tcolorbox}[colback=gray!10, colframe=black, boxrule=0.5pt, arc=1.5mm]
\begin{itemize}[nosep]
    \item[$\Box$] \textbf{Complete Verifiable (Hard Gate):} Is the subclaim a single complete sentence that makes one verifiable proposition?
    \item[$\Box$] \textbf{Retrieval Relevant (0.40):} Can a reader without access to the original question or context use this subclaim alone to retrieve correct, relevant evidence?
    \item[$\Box$] \textbf{Qualifiers Sufficient (0.15):} If the subclaim depends on qualifiers (time, attribution, conditions, scope, location, etc.), are they present?
    \item[$\Box$] \textbf{References Explicit (0.30):} Are entities, events, and relations explicitly named and unambiguous?
    \item[$\Box$] \textbf{No Ungrounded Additions (0.15):} Does the subclaim avoid adding information not present in the question, context, or sentence?
\end{itemize}
\end{tcolorbox}

We compute the checklist reward using Llama-3.3-70B-Instruct via Amazon Bedrock (temperature \(0.2\), top-p \(0.9\)). For each subclaim, the judge outputs \texttt{Yes}, \texttt{No}, or \texttt{NA} for each criterion. We apply a hard constraint: if \texttt{complete\_verifiable} \(\neq\) \texttt{Yes}, the subclaim receives zero reward. Otherwise, we compute a weighted average over the remaining non-\texttt{NA} criteria using weights: \(0.40\) (retrieval relevance), \(0.30\) (explicit references), \(0.15\) (qualifiers), and \(0.15\) (no ungrounded content). The final reward for a completion is the geometric mean of all subclaim scores, with invalid or malformed judge outputs assigned zero reward. The complete prompt is in Appendix~\ref{appendix:prompts}.

\subsection{Format Reward}
\begin{tcolorbox}[colback=gray!10, colframe=black, boxrule=0.5pt, arc=1.5mm]
\begin{itemize}[nosep]
    \item[$\Box$] \textbf{Tags Present (0.40):} Both \texttt{<think>...</think>} and \texttt{<output>...</output>} appear.
    \item[$\Box$] \textbf{Order \& Clean (0.10):} The completion contains only these two blocks in order, with nothing extraneous.
    \item[$\Box$] \textbf{List Parsed (0.40):} The \texttt{<output>} parses as a list of strings (strict JSON or lenient literal).
    \item[$\Box$] \textbf{List Quality (0.10):} All items are non-empty strings; at least one item is required.
\end{itemize}
\end{tcolorbox}

The format reward uses regex-based parsing to verify: (i) presence of \texttt{<think>} and \texttt{<output>} tags, (ii) correct tag ordering with no extraneous content, (iii) successful parsing of \texttt{<output>} as a list, and (iv) list validity (non-empty, no blank strings). We combine these as a weighted sum with weights \(0.40\) (tags present), \(0.10\) (ordering), \(0.40\) (parseable list), and \(0.10\) (valid list).

\section{Use of AI Assistants}
Portions of this manuscript benefited from the use of an AI assistant for language refinement. All editorial decisions and content oversight remained the sole responsibility of the authors.

%% file: sections/error_analysis.tex
Our work aims to enhance the decomposer for better verification accuracy. However, as our approach integrates multiple components, overall performance remains constrained by the limitations of retrieval and verification modules, which can still introduce errors despite improved decomposition. This is especially pronounced in open-world retrieval scenarios, such as web search, where temporal discrepancies in evidence—particularly for time-sensitive claims—can significantly impact results. To clarify the approach’s limitations, we conducted an error analysis of misclassified claims, identifying whether failures arose from decomposition quality, evidence insufficiency, annotation ambiguity, or verifier reasoning. Appendix Table~\ref{tab:error_analysis_blocks} shows representative error cases, illustrating the types of mismatches observed and their  causes.

One of the recurring patterns from the decomposition perspective was the model's tendency to over-support composite claims, in cases where claim decomposition led the model to treat complex, multi-part statements as a single unit, resulting in support based on partial evidence (Table~\ref{tab:error_analysis_blocks}, example 1).
Annotation ambiguity contributed to some mismatches, especially in cases where the ground truth label was debatable (Table~\ref{tab:error_analysis_blocks}, example 1). 
Verifier can lack world knowledge or overgeneralize. Verifier sometimes accepted ambiguous or outdated evidence, failing to critically assess its relevance or recency (Table~\ref{tab:error_analysis_blocks}, examples 2,3). 

%% file: sections/prompts.tex
\section{Prompts Used in Experiments}
\label{appendix:prompts}
In this section, we include all prompts used in our experiments.
\subsection{Decomposition System Prompt}
\begin{prompt}[title={Decomposition Prompt}]
You are an assistant for a group of fact-checkers. You will be given a question inside the
\texttt{<question></question>} tags, which was asked about a source text (it may also be referred to by other
names, e.g., a dataset). You will also be given an excerpt from a response to the question inside the
\texttt{<context></context>} tags. If the context contains ``[...]'', this means you are \textbf{not} seeing the full
response. You will also be given a target sentence from the response inside the
\texttt{<sentence></sentence>} tags.

Your task is to identify all specific and verifiable propositions in the target sentence and ensure each
proposition is \textbf{decontextualized} while maintaining any relationships between the claims. To do this, 
you must reason step by step wrapped inside \texttt{<think>...</think>} using the steps provided. The final results 
must always be output inside \texttt{<output>...</output>}. Follow the rules strictly.

\textbf{Follow these rules through the steps:}
\begin{itemize}
  \item A \textbf{verifiable proposition/claim} is a concrete, checkable assertion (event, state, relation,
  attribution) whose truth can be assessed against reliable external knowledge (e.g., Wikipedia or Google Search).
  The \textbf{truth} or \textbf{relevance} of the claim does \textbf{not} affect inclusion.
  \item Use the \textbf{Question/Context only} to resolve references (pronouns, definite phrases, acronyms),
  not external knowledge.
  \item Treat \textbf{attributions} as verifiable facts about speech/stance (``X says/claims/reports/
  predicts Y''), even if Y is speculative.
  \item \textbf{Include as verifiable}: entity/role facts; events (who/what/when/where); quantified facts
  (numbers/dates/comparisons); attributed statements/stances; context-bridged answers
  (e.g., Q: ``Who is CEO?'' → A: ``John.'').
  \item \textbf{Exclude}: lack-of-information statements; introductions/previews; pure conclusions/summaries;
  generic/normative opinions; unattributed speculation/hypotheticals; instructions/suggestions;
  headings/teasers.
  \item When a sentence mixes generic language with a specific fact, extract the \textbf{specific fact only}.
  \item \textbf{Decontextualized claims} must be self-contained, preserve the original meaning relative to
  context, and refer to entities by \textbf{name} (avoid pronouns). Include \textbf{time/place} when present.
  Keep each claim to one sentence with zero or at most one embedded clause.
  \item If Context contains ``[...]'', you may \textbf{not} assume unseen material.
\end{itemize}

\textbf{Follow these reasoning steps inside \texttt{<think></think>}:}

\textbf{STEP 1 — Decide if the sentence has verifiable claims (TargetSentence → VerifiableSentence)}\\
Input: Target sentence contained in \texttt{<sentence></sentence>}.\\
Determine whether this particular sentence contains at least one specific and verifiable proposition,
and if so, return a complete sentence that only contains verifiable information.\\
Apply Include/Exclude rules strictly.\\
If YES → produce VerifiableSentence with the verifiable information.\\
If NO → stop and final output is \texttt{<output>No verifiable claim</output>}.\\[2mm]

\textbf{STEP 2 — Rewrite the sentence so it is clear and self-contained (VerifiableSentence → DecontextualizedSentence)}\\
Input: VerifiableSentence from Step 1.\\
Replace pronouns, partial names, and undefined acronyms/abbreviations only if the full form appears in the Question or the Context.\\
If the sentence, when read in isolation, could have more than one meaning, use the Question/Context to select the correct interpretation so that the sentence can stand alone.\\
If ambiguity remains and multiple reasonable meanings cannot be resolved using the Question/Context, stop and output: \texttt{<output>Cannot be decontextualized</output>}.\\
Otherwise, produce: \texttt{DecontextualizedSentence:} a single, self-contained sentence that preserves the original meaning while resolving references and ambiguity using only the given Question/Context.\\[2mm]

\textbf{STEP 3 — Find relationships between the parts of the claim (DecontextualizedSentence → Relationships)}\\
Input: DecontextualizedSentence from Step 2. Identify relationships in the sentence, if present:
\begin{itemize}
  \item \textbf{Attribution} (e.g., ``X says Y.'')
  \item \textbf{Causal} (cause, effect, purpose, condition; e.g., ``X happened because Y.'' / ``If X, then Y.'')
  \item \textbf{Temporal} (time sequence or overlap; e.g., ``X happened after Y.'' / ``X happened while Y occurred.'')
  \item \textbf{Comparison} (contrast, concession, similarity; e.g., ``X increased, while Y decreased.'' / ``Even though X, Y happened.'')
  \item \textbf{Expansion} (elaboration, example, detail, alternative; e.g., ``X, such as Y.'' / ``Instead of X, Y.'')
  \item \textbf{Negation} (explicit negatives; e.g., ``X did not happen.'')
  \item \textbf{Membership/Part-of} (subset or composition; e.g., ``X is part of Y.'' / ``X is a member of Y.'')
\end{itemize}
Produce a list of identified relationships (or ``No relationships'').\\[2mm]

\textbf{STEP 4 — Split the claim into the smallest possible factual units (DecontextualizedSentence + Relationships → Claims)}\\
Input: DecontextualizedSentence from Step 2 and Relationships from Step 3.\\
Split the DecontextualizedSentence into \textbf{specific, verifiable, and fully decontextualized claims}, meaning it can be
understood in isolation (i.e., without the question, the context, and the other propositions), and its meaning in isolation matches its meaning when
interpreted alongside the question, the context, and the other propositions.\\
Preserve all \textbf{numbers, time, place, comparisons, negations, attributions}.\\
Do not invent details beyond the given text.\\
Quotations may be extracted verbatim but must be connected via \textbf{Attribution}.\\
Final output: \texttt{<output>[claim1, claim2, ...]</output>}.\\[2mm]

\textbf{Format the response as:}
\begin{verbatim}
<think>
  [Step 1: Selection reasoning → VerifiableSentence or stop 
  Step 2: Decontextualization reasoning → DecontextualizedSentence or stop  
  Step 3: Relationship identification reasoning → Relationships  
  Step 4: Decomposition reasoning → Claims]
</think>
<output>
  [Final result: either “No verifiable claim”, “Cannot be decontextualized”, 
   or list of decomposed claims from Step 4]
</output>
\end{verbatim}
\end{prompt}

\subsection{Decomposition Checklist Prompt}
\begin{prompt}[title={Decomposition Checklist Prompt}]
You are an assistant for a group of fact-checkers. Your job is to help verify whether the extracted subclaims from a previous claim extraction task satisfy various requirements for being
decontextualized—i.e., whether they stand alone and correctly encode the relationship asked by the checklist—using \textbf{ONLY} the materials provided. Do not use external knowledge. Do not approximate.

\textbf{INPUTS}\\
You will be given:\\
\texttt{<question>}the user’s question that generated the model’s response\texttt{</question}\\
\texttt{<context>}a portion of the model’s response surrounding the target sentence (including some preceding and following text)\texttt{</context}\\
\texttt{<sentence>}the specific sentence from the context to be decomposed\texttt{</sentence}\\
\texttt{<subclaims>["subclaim 1", "subclaim 2"]</subclaims>}

\textbf{NOTES}
\begin{itemize}[leftmargin=*, itemsep=2pt]
  \item Evaluate each subclaim independently; do not compare across subclaims.
  \item The \texttt{<context>} may be partial; if it contains \texttt{[...]} you are not seeing the full response.
  \item Judge only using information in \texttt{<question>}, \texttt{<context>}, \texttt{<sentence>}, and \texttt{<subclaims>}. Do not use outside knowledge.
  \item If something cannot be determined from the provided materials, mark the check as \texttt{NA}.
  \item Only checklist items \textbf{3} (\texttt{qualifiers\_sufficient}) and \textbf{4} (\texttt{references\_explicit}) may return \texttt{NA}.
\end{itemize}

\textbf{CHECKLIST}\\
For each subclaim, answer these questions with \texttt{Yes/No/NA}. Use the JSON keys exactly as shown.
\begin{enumerate}[leftmargin=*, itemsep=4pt]
  \item \textbf{complete\_verifiable}\\
  Is the subclaim a single complete sentence that makes one verifiable proposition?\\
  A verifiable proposition means a statement that can be objectively verified as true or false based on empirical evidence or reality (not an opinion, vague description, or evaluative language).
  \item \textbf{retrieval\_relevant}\\
  Can a reader with no access to the original question or context use this subclaim alone to retrieve the correct and relevant evidence for the sentence?\\
  \textit{Yes}: The subclaim is stand-alone and disambiguating—it explicitly names the target entity/event and includes the necessary salient identifiers from the provided materials (e.g., roles/titles, aliases, timeframe, locations, distinctive actions). It does not introduce unsupported attributes.\\
  \textit{No}: It is background that cannot be linked to the question, is off-topic, or too vague.
  \item \textbf{qualifiers\_sufficient}\\
  If the subclaim depends on qualifiers (time, attribution, conditions, scope, location, etc.) in order to be a complete and accurate statement, are those qualifiers present in the subclaim?\\
  If no qualifier is necessary for the subclaim to stand as a complete, faithful proposition, mark \texttt{Yes}.\\
  If it is unclear from the provided materials whether a qualifier is necessary, mark \texttt{NA}.
  \item \textbf{references\_explicit}\\
  Are all entities, events, and relations in the subclaim explicitly named and unambiguous, using only the information available in \texttt{<question>}, \texttt{<context>}, or \texttt{<sentence>}?\\
  Names must be fully resolved: if both a first and last name (or equivalent full form) are present in the materials, use the full form (e.g., ``John Smith'' instead of just ``John'' or ``Smith'').\\
  Pronouns are never acceptable: if the subclaim uses \textit{he, she, they, it, this, that, these, those, former, latter}, etc. in place of an entity or event → mark \texttt{No}.\\
  Acronyms must be expanded if the expansion is given in the materials; otherwise mark \texttt{No}.\\
  If ambiguity cannot be resolved from the provided materials → \texttt{NA}. Otherwise → \texttt{Yes}.
  \item \textbf{no\_ungrounded\_additions}\\
  Does the subclaim avoid adding information not explicitly present in the question, context, or sentence?
\end{enumerate}

\textbf{OUTPUT FORMAT}\\
Output \textbf{STRICT JSON only} (a top-level list of objects). Each object must include both \texttt{"checks"} and \texttt{"rationales"} (single-sentence, grounded in the provided materials, no external info).\\
Schema:
\texttt{\{}\texttt{[}\\
\hspace*{1em}\texttt{\{}\texttt{"id": "index",}\\
\hspace*{2em}\texttt{"checks": \{}\texttt{"complete\_verifiable": "Yes|No", "retrieval\_relevant": "Yes|No", "qualifiers\_sufficient": "Yes|No|NA", "references\_explicit": "Yes|No|NA", "no\_ungrounded\_additions": "Yes|No"\}\,,}\\
\hspace*{2em}\texttt{"rationales": \{}\texttt{"complete\_verifiable": "One concise sentence explaining the decision.", "retrieval\_relevant": "One concise sentence explaining the decision.", "qualifiers\_sufficient": "One concise sentence explaining the decision.", "references\_explicit": "One concise sentence explaining the decision.", "no\_ungrounded\_additions": "One concise sentence explaining the decision."\}}\\
\hspace*{1em}\texttt{\}}\texttt{]} \texttt{\}}

\medskip
\textbf{Here is your input:}\\
\texttt{<question>\{QUESTION\}</question>}\\
\texttt{<context>\{CONTEXT\}</context>}\\
\texttt{<sentence>\{SENTENCE\}</sentence>}\\
\texttt{<subclaims>\{SUBCLAIMS\_JSON\_ARRAY\}</subclaims>}
\end{prompt}